\documentclass[journal]{IEEEtran}
\IEEEoverridecommandlockouts
\ifCLASSINFOpdf
\else
\fi
\usepackage{multirow}
\usepackage{pdflscape}
\usepackage{amsmath,amssymb,amsfonts}
\usepackage{graphicx}
\usepackage{xcolor}
\usepackage{subfig}
\usepackage{mathtools}
\usepackage{cite}
\usepackage{array}
\usepackage{amsmath}
\usepackage{mathdots} 
\usepackage{amsthm}
\theoremstyle{plain}

\newtheorem{theorem}{Theorem}
\usepackage{float}

\def\BibTeX{{\rm B\kern-.05em{\sc i\kern-.025em b}\kern-.08em
    T\kern-.1667em\lower.7ex\hbox{E}\kern-.125emX}}

\begin{document}

\title{Super-resolution Multi-signal Direction-of-Arrival Estimation by Hankel-structured Sensing and Decomposition}

\author{Georgios I. Orfanidis,~\IEEEmembership{Student Member~IEEE}, ~Dimitris A. Pados,~\IEEEmembership{Senior Member~IEEE}, \\George Sklivanitis,~\IEEEmembership{Member~IEEE}, and~Elizabeth S. Bentley,~\IEEEmembership{Senior Member~IEEE}
\thanks{G. I. Orfanidis, D. A. Pados, and G. Sklivanitis are with the Center for Connected Autonomy and AI and the Dept. of Electrical Engineering and Computer Science, Florida Atlantic University, Boca Raton, FL 33431 USA (email: \{gorfanidis2021, dpados, gsklivanitis\}@fau.edu).

E. S. Bentley is with the Air Force Research Laboratory, AFRL/RI, Rome, NY 13441 USA (e-mail: elizabeth.bentley.3@us.af.mil).

This work was supported in part by the National Science Foundation under Grants CNS-2117822 and EEC-2133516 and the Air Force Research Laboratory under Grant FA8750-25-1-1000.
Distribution A. Approved for public release: Distribution Unlimited: AFRL-2026-2350 on 12 May 2026.}}
\maketitle

\begin{abstract}
 Motivated by sensing modalities in modern autonomous systems that involve hardware-constrained spatial sampling over large arrays with limited coherence time, we develop a novel framework for rapid super-resolution multi-signal direction-of-arrival (DoA) estimation based on Hankel-structured sensing and data matrix decomposition of arbitrary rank, under both the $L_2$ and $L_1$-norm formulation. The resulting $L_2$-norm estimator is shown to be maximum-likelihood optimal in white Gaussian noise. The $L_1$-norm estimator is shown to be maximum-likelihood optimal in independent, identically distributed (i.i.d.) isotropic Laplace noise, offering broad robustness to impulsive interference and corrupted measurements commonly encountered in practice. Extensive simulations demonstrate that the proposed methods exhibit powerful super-resolution capabilities, requiring significantly lower SNR and achieving substantially higher resolution probability than recent competing approaches.
\end{abstract}


\section{Introduction}
\IEEEPARstart {S}{ignal} Direction-of-Arrival (DoA) estimation has long been an embedded requirement in applications such as wireless communications, remote sensing, radar and sonar, to name a few \cite{van2002optimum,526899,622504,7815340,9219157,9725025,9722876}. Classical DoA estimation methods have been formulated within a statistical framework where reliable estimation requires the availability of sufficiently many antenna-array snapshots in order to estimate the underlying statistics of the received data. Maximum Likelihood (ML) approaches arise from directly optimizing the likelihood function associated with the observed snapshots  \cite{57542,scheweppe,17564}, whereas subspace (signal or noise) analysis methods exploit the eigenspace structure of the sample space-domain autocovariance matrix leading to the well-known methods of MUSIC, ESPRIT, etc. \cite{1143830,32276,80966,4063549,9516894}. 

The operational performance of ML-derived and subspace-based DoA estimators has been characterized asymptotically with respect to the number of antenna-array snapshots, the number of antenna-array elements, and the signal-to-noise ratio (SNR) \cite{17564}. While non-asymptotic performance (finite number of antenna-array elements, finite number of antenna-array snapshots, finite SNR values) is generally unknown, it has been established \cite{17564, 348129} that DoA accuracy is more adversely affected by small antenna-array size than small number of antenna-array snapshots. 
Today, there is real need for high performing, small-sample-support DoA estimation systems. The emergence of networked mobile autonomous systems operating in challenging environments (ground, air, space, underwater) and communicating over high-frequency bands (e.g., mm-wave, THz) has sparked interest in rapid and accurate signal DoA estimation estimation in order to sustain high-data-rate connectivity among cooperating agents and to mitigate interference from other in-band sources\cite{9148550,9726790,10160524,10018231}. Similar needs exist across a broad range of other contemporary applications, including real-time localization and tracking for Internet of Things (IoT) and smart city infrastructure \cite{8879484,9535488,10096647}, as well as emerging technologies such as mm-wave robotics \cite{9214481}, machine-to-machine communications \cite{7494995}, and automotive radar \cite{9362210}. Owing to the severely limited statistical coherence time in such scenarios, conventional DoA estimation techniques that rely on well estimated second-order or higher-order statistics are deemed inapplicable.


Motivated by such sensing environments, there has been growing interest in one-shot/few-shot DoA estimation. Recent works used approaches ranging from compressive sensing (CS) and deep learning (DL) to powerful linear algebraic advancement based on matrix decompositions. While many of these methods were originally developed for the extreme one-shot case, they naturally extend to the few-shot regime.
The emerging investigation of the CS framework in the context of DoA estimation 
stems from the observation that sparse signals can be reliably recovered from a limited number of antenna-array snapshots, as well as from the relatively low sensitivity of sparse reconstruction methods to SNR \cite{1468495,7744507}. Grid-based methods assume that the DoAs lie on a predefined angular grid, whereas grid-less methods operate directly in the
continuous angular domain. 
In grid-based formulations, the DoA estimation problem is typically cast as a sparse representation task using sparsity-promoting penalties, most commonly the $L_1$-norm, leading to convex programs such as LASSO and basis pursuit denoising (BPDN), along with related variants including square-root LASSO (SR-LASSO) and SPICE \cite{Belloni_2011,5599897,6553252, babu2014connection}. Extensions employing nonconvex $L_p$ penalties ($0<p<1$) (e.g., FOCUSS, SLIM) \cite{gorodnitsky1997sparse,rao2003subset,tan2010sparse}, direct $L_0$ approximations (SL$0$) \cite{4663911}, and weighted least-squares–based adaptive methods (IAA-APES) \cite{5417172} have also been proposed to enhance sparsity while attempting to reduce dependence on heuristic regularization parameters.

In contrast, grid-less sparse approaches treat DoA estimation as a continuous-domain line spectral estimation problem and promote sparsity via structured convex relaxations such as the atomic norm \cite{6576276} or Hankel-based nuclear norm \cite{6867345}. These formulations are typically expressed as semidefinite programs (SDPs) and solved using general-purpose convex optimization tools. While grid-less methods avoid discretization mismatch, they introduce additional design parameters—such as regularization variables or structured matrix dimensions, whose selection significantly impacts performance and computational complexity. The reader is referred to \cite{yang2018sparse,fortunati2014single} for comprehensive reviews and comparative analyses of CS-based DoA estimation frameworks.

While learning-based approaches to DoA estimation can be traced back to early neural network models such as Hopfield networks \cite{1169330,197061,120069}, recent advances in machine learning and their widespread adoption in signal processing \cite{10056957} renewed interest in applying deep neural networks (DNNs) to the DoA estimation problem. Motivated in part by the high computational complexity of classical maximum likelihood estimation, a variety of learning-based models have been proposed to enable fast inference.
Assuming prior knowledge of the number of sources, the work in \cite{8726554} introduced a multi-layer perceptron (MLP) that takes as input a transformed version of the normalized spatial covariance matrix and produces DoA estimates via post-processing the network's output, while \cite{9747692} proposed Deep-MLE, a hybrid framework that integrates an one-dimensional ResNet architecture with the conventional MLE formulation. In contrast, the approach in \cite{8682604} does not assume prior knowledge of the number of active sources and instead employs a neural network architecture that jointly estimates the number of sources and their corresponding DoAs by formulating the problem as classification and regression task, respectively. In an effort to design more interpretable and computationally efficient architectures, \cite{10348517} proposed Deep-MPDR, which embeds the minimum-power-distortionless-response (MPDR) beamformer within a deep learning framework.
Despite their fast inference capabilities, learning-based DoA estimators typically require extensive training on large and representative datasets, which are often difficult to obtain and curate to adequately capture real-world operating conditions. Moreover, the applicability of such estimators is generally restricted to scenarios that closely match the assumptions of the underlying training data, including the number of antenna elements, the number of active sources, and the operating SNR range, which may limit robustness in dynamically changing or hardware-constrained environments.

In few-shot DoA estimation settings, where only a limited number of antenna-array snapshots are available, the sample space-domain autocorrelation matrix is often poorly conditioned and may become severely rank deficient. As a result, conventional subspace-based DoA estimation methods become ill-defined or unreliable. To address these limitations, several works have leveraged developments in linear algebra and structured matrix estimation \cite{10051870,orfanidis2022time}. In particular, these approaches exploit the shift-invariance properties inherent in array measurements by embedding the received samples into structured matrices, most commonly of Hankel or Toeplitz form.
In the extreme case of one-shot DoA estimation, \cite{9965430, degen2017single} proposed constructing an estimate of the autocorrelation matrix from a single array snapshot, thereby restoring full rank and enabling the application of subspace-based estimators. However, the approach in \cite{degen2017single} is inherently limited to array architectures whose elements share identical statistical properties, such as uniform linear arrays (ULAs). Another representative structured-matrix approach is the Matrix Pencil method, originally developed for pole estimation in linear systems \cite{hua2002matrix} and later adapted for one-shot DoA estimation \cite{adve1996elimination} by forming a Hankel matrix from a single array snapshot. Similarly, \cite{liao2016music} proposed a single-snapshot variant of MUSIC in which the signal and noise subspaces are obtained from the singular value decomposition (SVD) of a Hankel matrix constructed from the observed measurement. Another widely used technique is spatial smoothing \cite{1164649,9564893}, which partitions the antenna array into overlapping subarrays and forms a spatially smoothed covariance matrix by averaging the subarray covariance estimates. This preprocessing step restores the rank of the covariance matrix and enables the application of classical subspace-based DoA estimators even in data-limited scenarios. Notably, the overlapping subarray partitioning inherent in spatial smoothing implicitly induces a Hankel-type structure in the resulting data representation. 


Today, with the proliferation of massive antenna arrays,  there is high hope for high-performing, high-resolution DoA estimation from limited antenna-array snapshots. There is, however, a complication that cannot be disregarded. Practical large antenna-array implementations commonly employ RF-switched, multiplexed, or low-power front-end designs in which the number of available RF chains, analog-to-digital converters, and processing resources is strictly limited. As a result, simultaneous observation of the full antenna-array aperture is often infeasible. 
In this paper, we consider the problem of estimating the direction of arrival of multiple signals with an antenna array of size larger than the number of implemented RF chains and we make the following new contributions:
\begin{itemize}
\item We propose sliding-window RF-chain sensing from one end of the array to the other. When sliding is done one element at a time to complete one run over the array, we call the process \emph{Hankel sensing}. Appending the sliding-window measurements side-by-side creates our data matrix with rows as many as the available RF chains and columns as many as the slides.

\item We propose two new DoA estimation algorithms based on rank-$K$ Hankel-structured decomposition of the data matrix formed above, under the $L_2$-norm and the $L_1$-norm formulation. 

\item We prove that $L_2$-norm rank-$K$ Hankel-structured decomposition produces jointly ML-optimal DoA estimates for the case of $K$ signals in white Gaussian noise. 

\item We prove that $L_1$-norm rank-$K$ Hankel-structured decomposition produces jointly ML-optimal DoA estimates for the case of $K$ signals in i.i.d. isotropic Laplace noise. 

    
 \item We carry out extensive simulation studies to evaluate the performance and resolution capabilities of the proposed estimators across varying signal-to-noise ratio (SNR) levels and angular separations between impinging sources. The evaluation considers both additive white Gaussian noise and Bernoulli–Gaussian mixture noise models, the latter emulating challenging environments with impulsive disturbances that may arise from potential hardware malfunctions.

\end{itemize}

The remainder of the paper is organized as follows. Section II introduces the signal model and the proposed sensing architecture, along with the necessary notation and technical preliminaries. Section III presents the proposed DoA estimation framework based on rank-$K$ Hankel-structured decompositions under both the $L_2$ and $L_1$-norm formulation. Section IV provides extensive simulation studies and comparative evaluations under both Gaussian and impulsive noise environments. Finally, conclusions are drawn in Section V.

\textit{Notation}: In this paper, matrices are denoted by upper-case bold letters, column vectors by lower-case bold letters, and scalars by lower-case plain-font letters. The transpose operation is represented by the superscript $(\cdot)^T$, conjugation by $(\cdot)^*$, the conjugate transpose (Hermitian) by $(\cdot)^H$, and the Kronecker product by $\otimes$.

\section{Signal Model and Hankel-sensing Architecture}
\label{signal&sensing_model}

For clarity and simplicity in presentation, we consider a uniform linear antenna array (ULA) consisting of $M$ elements. Let $\theta \in [-90^{\circ}, 90^{\circ})$ represent the direction of arrival of an impinging signal measured with respect to the array broadside. The corresponding array response vector is given by
\begin{equation}
\label{array_response_vector}
\mathbf{a}(\theta)
\triangleq
\left[
1,\,
e^{-j 2 \pi \frac{d}{\lambda} \sin \theta},\,
\ldots,\,
e^{-j 2 \pi (M-1)\frac{d}{\lambda} \sin \theta}
\right]^T 
\end{equation}
where $d$ denotes the inter-element spacing and $\lambda$ is the carrier wavelength. Suppose that $K<M$ narrowband far-field sources of interest impinge on  the array from distinct directions $\theta_1,\ldots,\theta_K$. The baseband measurement collected across the $M$ sensors is expressed in vector form by
\begin{equation}
\label{signal_model_K_signals}
\mathbf{r}=\sum_{k=1}^{K} x_k \mathbf{a}(\theta_k)+\mathbf{n}
\end{equation}
where $x_k \in \mathbb{C}$ denotes the complex fixed unknown signal amplitude of the $k$th source and $\mathbf{n} \in \mathbb{C}^M$ represents additive measurement noise  with independent and identically distributed (i.i.d.) real and imaginary parts across sensors.

\begin{figure}[htbp]
    \centering
    \includegraphics[width=0.8\linewidth]{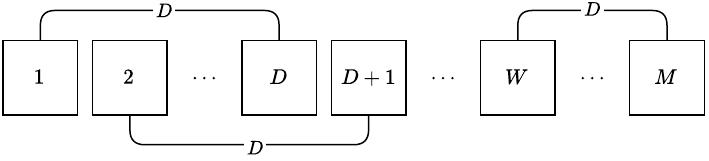}
    \caption{Sliding-window sensing across a uniform linear array of $M$ antenna elements using overlapping subarrays of length $D$. A total of $W=M-D+1$ subarray measurements are obtained by sequentially shifting the sensing window across the array.}
    \label{fig:sliding_array_arch}
\end{figure}

In many practical large antenna-array implementations, due to hardware constraints such as limited number of RF chains, power consumption considerations, or restrictions on the number of available analog-to-digital converters, it is not possible to simultaneously acquire measurements from all $M$ sensors.
Motivated by such switched sensing architectures, we propose a sequential data acquisition strategy in which measurements are collected from overlapping subarrays of the physical array. Specifically, as illustrated in Fig. \ref{fig:sliding_array_arch}, a sensing window of length $D$ (corresponding to the number of available RF chains) is slid across the array one element at a time. At each position of the window, a $D$-dimensional subarray observation is obtained. By shifting the window one sensor at a time, a total of $W = M-D+1$ overlapping subarray measurements are collected, collectively spanning the full physical aperture of the array. 

The observation corresponding to the $i$th subarray, $i = 0,\ldots,W-1$, can be written as
\begin{equation}
\label{sliding_sub}
\mathbf{r}_i
=
\left[
\sum_{k=1}^{K} x_k \mathbf{a}_{i+1}(\theta_k),
\ldots,
\sum_{k=1}^{K} x_k \mathbf{a}_{i+D}(\theta_k)
\right]^T
+
\mathbf{n}_i \\
\end{equation}
where $\mathbf{a}_j(\theta_k)$ denotes the $j$th entry of the response vector associated with source $k$ and $\mathbf{n}_i \in \mathbb{C}^{D}$ represents the additive noise vector corresponding to the $i$th subarray acquisition.

Next, all subarray observations are collected and arranged column-wise to form the aggregated measurement matrix $\mathbf{X} \in \mathbb{C}^{D \times W}$,
\begin{equation}
\label{agg_meas_matrix}
\mathbf{X} = [\mathbf{r}_0,\mathbf{r}_1,\ldots,\mathbf{r}_{W-1}].
\end{equation}
The sensing architecture under consideration induces a structured 
representation of the signal component in the aggregated $\mathbf{X}$.
In particular, the signal component in $\mathbf{X}$ admits a $K$-rank Hankel\footnote{A matrix is called Hankel if each anti-diagonal has elements of constant value \cite{golub1996matrix}.} representation.

\section{Proposed Multiple-Signals DoA Estimators}
\label{rank_k_Hankel_decomp}

\subsection{Rank-$K$ Hankel Decompositions and Multiple-Signal Direction-of-Arrival Estimation}


We know  that a matrix  $\mathbf{A} \in \mathbb{C}^{D \times W}$ is Hankel of rank one if and only if  $\mathbf{A}$ can be written as $\mathbf{A} = \frac{c}{\|\tilde{\mathbf{s}}_D(z)\| \, \|\tilde{\mathbf{s}}_W(z)\|} \, \tilde{\mathbf{s}}_D(z) \tilde{\mathbf{s}}_W(z)^T$ where $c \in \mathbb{C}\setminus\{0\}$,
\begin{equation}
    \tilde{\mathbf{s}}_D(z) = [1, z, z^2, \ldots, z^{D-1}]^{T},
\end{equation}
\begin{equation}
    \tilde{\mathbf{s}}_W(z) = [1, z, z^2, \ldots, z^{W-1}]^{T},
\end{equation}
$z \in \mathbb{C} \cup\{\infty\}$ \cite{knirsch2021optimal}. Recent work in \cite{orfanidis2026hankel} developed practical algorithms for rank-1 Hankel-structured decomposition of arbitrary complex matrices and applied these algorithms to the problem of single-signal DoA estimation.

Here, we focus on the important generalized problem of estimating jointly multiple signal DoAs. An underlying mathematical complication to overcome is that, in general, a complex matrix 
$\mathbf{A} \in \mathbb{C}^{D \times W}$ can be Hankel of rank-$K$, $K>1$, and still be written as a sum of lower rank matrices that are not necessary Hankel (addition of non-Hankel matrices can produce a Hankel matrix). 

Toward building an algorithm for joint DoA estimation of multiple signals from $\mathbf{X}$, we first need to normalize the polynomial vector representation used across signals to
\begin{equation}
\label{structure_vector}
{\mathbf{s}}_D(z) \triangleq \frac{\tilde{\mathbf{s}}_D(z)}{\left\|{\tilde{\mathbf{s}}_D(z)}\right\|_2}
\end{equation}
where, in closed form,
\begin{equation}
\label{norm}
\left\|{\mathbf{s}}_D(z)\right\|_2 =
\begin{cases}
\sqrt{\frac{1-|z|^{2D}}{1-|z|^2}}, & |z| \neq 1, \\
\sqrt{D}, & |z|=1 .
\end{cases}
\end{equation}
Under normalization, the generic problem of approximating an arbitrary complex matrix by a rank-$K$ Hankel matrix that consists of $K$ individual rank-$1$ components 
takes the form
\begin{equation}
\label{rankK_param}
(\hat{\mathbf{c}},\hat{\mathbf{z}}) = \underset{\substack{\mathbf{c}\in\mathbb{C}^K, \,
\mathbf{z}\in\mathbb{C}^K}} {\operatorname{argmin}} \left\| \mathbf{X}-\sum_{k=1}^{K} c_k\,{\mathbf{s}}_D(z_k){\mathbf{s}}_W(z_k)^T \right\|
\end{equation}
where $\mathbf{c}=[c_1,\ldots,c_K]^T$, $\mathbf{z}=[z_1,\ldots,z_K]^T$, and $|| \cdot ||$ is an approximation norm of choice. 
%
Parameterizing $z_k$ in the form of
\begin{equation}
    z(\theta_k) \triangleq e^{-j 2 \pi \frac{d}{\lambda} \sin \theta_k}, 
    \quad \theta_k \in [-90^\circ,90^\circ),
\end{equation} 
we transfer the search for the optimal $z_k$ value directly into the angular domain of interest. Then, the problem in (\ref{rankK_param}) is rewritten as 
\begin{equation}
\label{rankK_param_theta}
(\hat{\mathbf{c}},\hat{\boldsymbol{\theta}})
=
\underset{\substack{\mathbf{c} \in \mathbb{C}^{K}, \\ \boldsymbol{\theta} \in\left[-90^{\circ}, 90^{\circ}\right)^K}}
{\operatorname{argmin}}
\left\|
\mathbf{X}-\sum_{k=1}^{K} c_k\,{\mathbf{s}}_D(z(\theta_k)){\mathbf{s}}_W(z(\theta_k))^T
\right\|
\end{equation}
where $\boldsymbol{\theta}=[\theta_1,\ldots,\theta_K]^T$ denotes the vector collecting the corresponding signal directions. Therefore, irrespective of the particular matrix norm employed, the search over the complex parameters $z_k$ is restricted to the unit circle, significantly reducing the computational burden.

Finally, we note that the proposed formulation is not restricted only to sensing architectures whose aggregated measurement matrix exhibits a signal component of Hankel structure. In particular, certain array acquisition mechanisms can produce aggregated matrices with Toeplitz signal structure. Toeplitz and Hankel matrices are related through a simple reversal transformation by the anti-diagonal unit matrix $\mathbf{J}_D$.
The Toeplitz approximation problem for an arbitrary matrix $\mathbf{X}\in\mathbb{C}^{D\times W}$ can be equivalently expressed as a Hankel approximation problem applied to the transformed matrix $\mathbf{J}_D\mathbf{X}$ \cite{knirsch2021optimal}. The equivalence holds for all matrix norms that are invariant under unitary or permutation transformations, such as those induced by $\mathbf{J}_D$. Since $\mathbf{J}_D$ is invertible, multiplication by $\mathbf{J}_D$ preserves matrix rank.

Below, we derive practical estimators for $K$-signal DoA estimation based on rank-$K$ Hankel decompositions. Specifically, we pursue two different formulations: \emph{(i)} $L_2$-norm formulation, which yields efficient least-squares estimators, and \emph{(ii)} $L_1$-norm formulation, which yields least-absolute-deviation estimators and provides robustness to impulsive disturbances and heavy-tailed noise.

\subsection{$L_2$-norm $K$-signal DoA Estimation}
Under the $L_2$-norm formulation the problem in (\ref{rankK_param_theta}) becomes
\begin{IEEEeqnarray}{rCl}
\IEEEeqnarraymulticol{3}{l}{
(\hat{\mathbf{c}}_{L_2},\hat{\boldsymbol{\theta}}_{L_2})=
} \nonumber\\[4pt]
&& \displaystyle \underset{\substack{\mathbf{c} \in \mathbb{C}^{K}, \\ \boldsymbol{\theta} \in\left[-90^{\circ}, 90^{\circ}\right)^K}} {\operatorname{argmin}}
\left\| \mathbf{X}-\sum_{k=1}^{K} c_k\,\mathbf{s}_D(z(\theta_k))\mathbf{s}_W(z(\theta_k))^T \right\|_2 \nonumber\\[3pt]
\label{l2}
\end{IEEEeqnarray}
where $\|\cdot\|_2$ denotes the element-wise (Frobenius) $L_2$ matrix norm. In vectorized form we rewrite 
\begin{equation}
\label{l2_vec}
    (\hat{\mathbf{c}}_{L_2},\hat{\boldsymbol{\theta}}_{L_2})= \displaystyle \underset{\substack{\mathbf{c} \in \mathbb{C}^{K}, \\ \boldsymbol{\theta} \in\left[-90^{\circ}, 90^{\circ}\right)^K}} {\operatorname{argmin}}\left\| \mathrm{vec}(\mathbf{X})- \mathbf{S}(\boldsymbol{\theta})\mathbf{c}\right\|_2
\end{equation} where 
\begin{equation}
\label{s_theta}
    \mathbf{S}(\boldsymbol{\theta}) \triangleq \left[ \mathbf{s}_W(z(\theta_1)) \otimes \mathbf{s}_D(z(\theta_1)), \cdots, \mathbf{s}_W(z(\theta_K)) \otimes \mathbf{s}_D(z(\theta_K)) \right]
\end{equation} represents the ${DW \times K}$ Hankel structure manifold matrix. This is a linear least-squares problem with respect to $\mathbf{c}$ for fixed $\boldsymbol{\theta}$. Provided that the matrix $\mathbf{S}(\boldsymbol{\theta})$ has full column rank, the optimal coefficient vector admits the closed-form least-squares solution
\begin{equation}
\label{c_l2}
    \hat{\mathbf{c}}_{L_2}(\boldsymbol{\theta}) = \bigl(\mathbf{S}(\boldsymbol{\theta})^H \mathbf{S}(\boldsymbol{\theta})\bigl)^{-1}\mathbf{S}(\boldsymbol{\theta})^H\mathrm{vec}(\mathbf{X}).
\end{equation}
The full column rank condition holds whenever the steering vectors $\mathbf{s}_W(\theta_k) \otimes \mathbf{s}_D(\theta_k),$ $k=1,\cdots,K$,
are linearly independent, which is satisfied for distinct directions
$\theta_k$ and when $K \le DW$. Under this condition, substituting
$\hat{\mathbf{c}}_{L_2}(\boldsymbol{\theta})$ into the objective
function yields an optimization problem that depends only on the
direction parameters $\boldsymbol{\theta}$,
\begin{equation}
\label{l2_vec_theta}
    \hat{\boldsymbol{\theta}}_{L_2}= \displaystyle \underset{\substack{ \boldsymbol{\theta} \in\left[-90^{\circ}, 90^{\circ}\right)^K}} {\operatorname{argmin}}\left\| \mathrm{vec}(\mathbf{X})- \mathbf{S}(\boldsymbol{\theta})\hat{\mathbf{c}}_{L_2}(\boldsymbol{\theta})\right\|_2.
\end{equation}
The complete procedure for the proposed $L_2$-norm $K$-signal DoA estimator is summarized in Fig. \ref{l2_doa_alg}.
\begin{figure}[htbp]
\renewcommand{\arraystretch}{1.15}
{\fontsize{9pt}{9}\selectfont
    {\hrule height 0.2mm}
    \vspace{0.3mm}
    {\hrule height 0.2mm}
    \vspace{1mm}
    {\bf $L_2$-norm $K$-signal DoA  Estimation from Hankel-sensed Data}
    \vspace{0.5mm}
    {\hrule height 0.2mm}
    \vspace{2mm}
    \textbf{Input: } Hankel-sensed data matrix $\mathbf{X} \in \mathbb{C}^{D \times W}$; targeted number of signals $K$. \\[1mm]
    \begin{tabular}{r l}
        1: & Parametrize: $z(\theta_k) \triangleq e^{-j 2 \pi \frac{d}{\lambda} \sin \theta_k}, \quad \theta_k \in [-90^\circ,90^\circ)$, \\[1mm]
        & $k=1, \ldots, K$.\\[1mm]
        2: & Set: \\[1mm]
        & $\mathbf{S}(\boldsymbol{\theta}) \triangleq \left[ \mathbf{s}_W(z(\theta_1)) \otimes \mathbf{s}_D(z(\theta_1)), \cdots, \mathbf{s}_W(z(\theta_K)) \otimes \mathbf{s}_D(z(\theta_K)) \right]$.
        \\[1mm]
        3: & Compute: 
        $ \hat{\mathbf{c}}_{L_2}(\boldsymbol{\theta}) = \bigl(\mathbf{S}(\boldsymbol{\theta})^H \mathbf{S}(\boldsymbol{\theta})\bigl)^{-1}\mathbf{S}(\boldsymbol{\theta})^H\mathrm{vec}(\mathbf{X}).$ \\[1mm]
        4: & Solve: $\hat{\boldsymbol{\theta}}_{L_2}= \displaystyle \underset{\substack{ \boldsymbol{\theta} \in\left[-90^{\circ}, 90^{\circ}\right)^K}} {\operatorname{argmin}}\left\| \mathrm{vec}(\mathbf{X})- \mathbf{S}(\boldsymbol{\theta})\hat{\mathbf{c}}_{L_2}(\boldsymbol{\theta})\right\|_2$. 
    \end{tabular} 
    \textbf{Output:} $\hat{\boldsymbol{\theta}}_{L_2}$ estimate of $K$ DoAs. 
    \\[1mm]
    {\hrule height 0.2mm}
    \vspace{0.3mm}
    {\hrule height 0.2mm}
    \vspace{0.4cm}
}
\caption{Summary of proposed $L_2$-norm $K$-signal DoA estimation algorithm from Hankel-sensed data.}
\vspace{-0.2cm}
\label{l2_doa_alg}
\renewcommand{\arraystretch}{1}
\end{figure}

The following theorem establishes the maximum-likelihood optimality of the proposed $L_2$-norm $K$-signal DoA estimator from the data matrix $\mathbf{X}$ in the presence of i.i.d white Gaussian noise.
The proof is given in the Appendix.
\begin{theorem}
\label{l2_ml_equivalence}
Under independent identically distributed (i.i.d.) circularly symmetric complex Gaussian noise per sensor, the proposed estimator in (\ref{l2_vec_theta}) is maximum-likelihood optimal. \hfill $\blacksquare$
\end{theorem}

\subsection{$L_1$-norm $K$-signal DoA Estimation}
Under the $L_1$-norm formulation the problem in (\ref{rankK_param_theta}) becomes
\begin{IEEEeqnarray}{rCl}
\IEEEeqnarraymulticol{3}{l}{
(\hat{\mathbf{c}}_{L_1},\hat{\boldsymbol{\theta}}_{L_1})=
} \nonumber\\[4pt]
&& \displaystyle \underset{\substack{\mathbf{c} \in \mathbb{C}^{K}, \\ \boldsymbol{\theta} \in\left[-90^{\circ}, 90^{\circ}\right)^K}} {\operatorname{argmin}}
\left\| \mathbf{X}-\sum_{k=1}^{K} c_k\,\mathbf{s}_D(z(\theta_k))\mathbf{s}_W(z(\theta_k))^T \right\|_1 \nonumber\\[3pt]
\label{l1}
\end{IEEEeqnarray}
where $\|\cdot\|_1$ denotes the element-wise $L_1$ matrix norm. In vectorized form we rewrite 
\begin{equation}
\label{l1_vec}
    (\hat{\mathbf{c}}_{L_1},\hat{\boldsymbol{\theta}}_{L_1})= \displaystyle \underset{\substack{\mathbf{c} \in \mathbb{C}^{K}, \\ \boldsymbol{\theta} \in\left[-90^{\circ}, 90^{\circ}\right)^K}} {\operatorname{argmin}}\left\| \mathrm{vec}(\mathbf{X})- \mathbf{S}(\boldsymbol{\theta})\mathbf{c}\right\|_1.
\end{equation} 
This is a
least-absolute-deviations (LAD) problem with respect to
$\mathbf{c}$ for fixed $\boldsymbol{\theta}$ \cite{birkes2011alternative,bloomfield1980least}. In contrast to the
$L_2$-norm formulation, which results in a linear least-squares
problem admitting a closed-form solution, the $L_1$-norm criterion instead, must be solved numerically using convex optimization
techniques. Specifically, for a fixed $\boldsymbol{\theta}$ the coefficient vector $\mathbf{c}$ is obtained as the solution of
\begin{equation}
\label{lad}
    \hat{\mathbf{c}}_{L_1}(\boldsymbol{\theta}) \in \displaystyle \underset{\substack{\mathbf{c} \in \mathbb{C}^{K}}} {\operatorname{argmin}}\left\| \mathrm{vec}(\mathbf{X})- \mathbf{S}(\boldsymbol{\theta})\mathbf{c}\right\|_1.
\end{equation} 
This problem is convex and can be equivalently reformulated as a linear programming (LP) problem by introducing auxiliary variables that bound the absolute residuals, thereby enabling efficient numerical solution using standard LP solvers \cite{boyd2004convex}. Alternatively, it can be expressed as a second-order cone program (SOCP) and solved using interior-point methods within modern convex-optimization frameworks \cite{boyd2004convex}. In addition to these general-purpose approaches, specialized algorithms for $L_1$-norm minimization can also be employed, including iteratively reweighted least squares (IRLS) schemes \cite{schlossmacher1973} and subgradient-based optimization methods, which provide computationally efficient means for obtaining the LAD solution \cite{wesolowsky1981}.

Consequently, the overall DoA estimation problem reduces to optimization over the direction parameters $\boldsymbol{\theta}$ only,
\begin{equation}
\label{l1_vec_theta}
    \hat{\boldsymbol{\theta}}_{L_1}= \displaystyle \underset{\substack{ \boldsymbol{\theta} \in\left[-90^{\circ}, 90^{\circ}\right)^K}} {\operatorname{argmin}}\left\| \mathrm{vec}(\mathbf{X})- \mathbf{S}(\boldsymbol{\theta})\hat{\mathbf{c}}_{L_1}(\boldsymbol{\theta})\right\|_1
\end{equation} where $\hat{\mathbf{c}}_{L_1}(\boldsymbol{\theta})$ is given in (\ref{lad}). 

The complete procedure for the proposed $L_1$-norm $K$-signal DoA estimator is summarized in  Fig. \ref{l1_doa_alg}. The theorem that follows 
establishes the maximum-likelihood optimality of the proposed $L_1$-norm $K$-signal DoA estimator from the data matrix $\mathbf{X}$ in the presence of symmetric complex Laplace noise. The proof is given in the Appendix.

\begin{theorem}
\label{ml_laplace}
Under i.i.d. circularly symmetric complex Laplace noise per sensor, the proposed estimator in (\ref{l1_vec_theta}) is maximum-likelihood optimal. \hfill $\blacksquare$
\end{theorem}

\begin{figure}[htbp]
\renewcommand{\arraystretch}{1.15}
{\fontsize{9pt}{9}\selectfont
    {\hrule height 0.2mm}
    \vspace{0.3mm}
    {\hrule height 0.2mm}
    \vspace{1mm}
    {\bf $L_1$-norm $K$-signal DoA  Estimation from Hankel-sensed Data}
    \vspace{0.5mm}
    {\hrule height 0.2mm}
    \vspace{2mm}
    \textbf{Input: } Hankel-sensed data matrix $\mathbf{X} \in \mathbb{C}^{D \times W}$; targeted number of signals $K$. \\[1mm]
    \begin{tabular}{r l}
        1: & Parametrize: $z(\theta_k) \triangleq e^{-j 2 \pi \frac{d}{\lambda} \sin \theta_k}, \quad \theta_k \in [-90^\circ,90^\circ)$, \\[1mm]
        & $k=1, \ldots, K$.\\[1mm]
        2: & Set: \\[1mm]
        & $\mathbf{S}(\boldsymbol{\theta}) \triangleq \left[ \mathbf{s}_W(z(\theta_1)) \otimes \mathbf{s}_D(z(\theta_1)), \cdots, \mathbf{s}_W(z(\theta_K)) \otimes \mathbf{s}_D(z(\theta_K)) \right]$.
        \\[1mm]
        3: & Compute: 
        $
        \hat{\mathbf{c}}_{L_1}(\boldsymbol{\theta}) \in \displaystyle \underset{\substack{\mathbf{c} \in \mathbb{C}^{K}}} {\operatorname{argmin}}\left\| \mathrm{vec}(\mathbf{X})- \mathbf{S}(\boldsymbol{\theta})\mathbf{c}\right\|_1.$\\[1mm]
        4: & Solve: $\hat{\boldsymbol{\theta}}_{L_1}= \displaystyle \underset{\substack{ \boldsymbol{\theta} \in\left[-90^{\circ}, 90^{\circ}\right)^K}} {\operatorname{argmin}}\left\| \mathrm{vec}(\mathbf{X})- \mathbf{S}(\boldsymbol{\theta})\hat{\mathbf{c}}_{L_1}(\boldsymbol{\theta})\right\|_1$. 
    \end{tabular} 
    \textbf{Output:} $\hat{\boldsymbol{\theta}}_{L_1}$ estimate of $K$ DoAs. 
    \\[1mm]
    {\hrule height 0.2mm}
    \vspace{0.3mm}
    {\hrule height 0.2mm}
    \vspace{0.4cm}
}
\caption{Summary of proposed $L_1$-norm $K$-signal DoA estimation algorithm from Hankel-sensed data.}
\vspace{-0.2cm}
\label{l1_doa_alg}
\renewcommand{\arraystretch}{1}
\end{figure}

\section{Simulation studies and comparisons}

In this section, we present simulation studies designed to evaluate the performance of the proposed DoA estimators under the sensing architecture introduced in Section \ref{signal&sensing_model}. The experiments focus on assessing the resolution capabilities of the proposed Hankel matrix decomposition–based estimators under both the $L_2$ and $L_1$-norm formulations in the presence of white Gaussian noise as well as impulsive noise that emulates spectral disturbances and hardware imperfections. For varying number of antenna-array elements, we investigate several performance metrics commonly used in array processing, including the expected signal-to-noise ratio (SNR) required to resolve closely spaced sources and the probability of resolution as a function of both the SNR and the angular separation $\Delta\theta$.

In all simulations, we consider a uniform linear array (ULA) with Nyquist inter-element spacing $d=\frac{\lambda}{2}$. We assume the availability of $D=\frac{M}{2}$ RF processing chains and employ a predefined angular search step over the angle-of-arrival domain of $0.25^\circ$ for all methods under evaluation. 
The experiments involve two sources of equal power located at directions $\theta_1$ and $\theta_2$. Without loss of generality, the two sources are placed symmetrically around the broadside such that
\[
\theta_1=-\frac{\Delta\theta}{2} \quad \text{and} \quad \theta_2=\frac{\Delta\theta}{2}
\] where $\Delta\theta = |\theta_1-\theta_2|$ denotes the angular separation between the two sources. Following the resolution criterion in \cite{van2002optimum}, the two sources are declared to be successfully resolved if and only if their estimated directions $\hat{\theta}_1$ and $\hat{\theta}_2$ satisfy
\begin{equation}
    |\theta_1-\hat{\theta}_1| < \frac{\Delta\theta}{2}
    \quad \text{and} \quad
    |\theta_2-\hat{\theta}_2| < \frac{\Delta\theta}{2}.
\end{equation}

For benchmarking, we compare the proposed estimators with several representative DoA estimation methods from the literature, commonly employed in small-sample-support scenarios. 
In particular, we consider \emph{(i)} the Matrix Pencil (\emph{``MP''}) method \cite{adve1996elimination} and \emph{(ii)} the single-snapshot MUSIC (\emph{``one-shot MUSIC''}) approach \cite{liao2016music}, both of which estimate signal parameters through Hankel embedding of the received measurement. We also include spatial smoothing–based estimators \cite{1164649,9564893} (e.g., Forward-Backward), which partition the antenna array into overlapping subarrays and estimate a spatially smoothed covariance matrix by averaging the resulting subarray covariances. 
Notably, the overlapping subarray structure induced by spatial smoothing yields an implicit Hankel-type representation and closely resembles the sensing architecture considered in this work, making it a particularly relevant baseline for comparison. In this framework, we evaluate the performance of classical estimators, namely MUSIC and MVDR, applied to the resulting covariance matrix, referred to as \emph{(iii) ``FBSS MUSIC''} and \emph{(iv) ``FBSS MVDR'',} respectively.
To further benchmark against methods designed for conventional sensing architectures that simultaneously observe all antenna elements, 
repeated antenna-array element measurements introduced by the sliding-window sensing architecture are averaged according to their multiplicities to form the full-aperture array snapshot. Based on this reconstruction, \emph{(v)} we consider the Matched-Filter (MF) estimator applied to the Multiplicity-Averaged (MA) snapshot, referred to as \emph{``MA-MF'',} which evaluates the energy of the data vector projected onto the subspace spanned by the steering vectors. It is worth noting that projection energy is the joint deterministic maximum-likelihood estimator in the single-snapshot setting for white Gaussian noise \cite{526899,hacker2010single}. However, operating on the multiplicity-averaged data the estimator does not correspond to the true ML solution any more under the Hankel sensing model, as established by Theorem 1. In addition, we consider the approach proposed in \cite{degen2017single}, which addresses the rank deficiency of the single-snapshot covariance matrix via structured covariance reconstruction. Specifically, exploiting a Toeplitz structure \cite{marple2019digital}, a consistent covariance estimate is obtained, thereby enabling the application of classical subspace-based DoA estimators such as MUSIC and MVDR. In the sequel, this approach is referred to as \emph{(vi)} multiplicity-averaged Toeplitz covariance MUSIC (\emph{``MA Toeplitz MUSIC''}) and \emph{(vii)} MVDR (\emph{``MA Toeplitz MVDR''}), respectively.

We first consider the additive white Gaussian noise (AWGN) disturbance model. We recall that under the assumption of white Gaussian noise the proposed DoA estimator based on rank-$K$ Hankel decompositions with the $L_2$-norm (Fig. \ref{l2_doa_alg}) is ML optimal as established in Theorem \ref{l2_ml_equivalence}. In the simulations, the signal-to-noise ratio (SNR), defined as $\mathrm{SNR} \triangleq \frac{|x|^2}{\sigma^2}$ in dB for noise variance $\sigma^2$, is controlled by setting the signal amplitude to $x=\sqrt{10^{\mathrm{SNR}/10}\sigma^2}\,e^{j\phi}$ where $\phi$ is an arbitrary phase drawn uniformly from $[0,2\pi]$.

\begin{figure}[htbp]
    \centering
    \subfloat[]{
        \label{fig:l2_M_16}
        \includegraphics[width=0.48\textwidth]{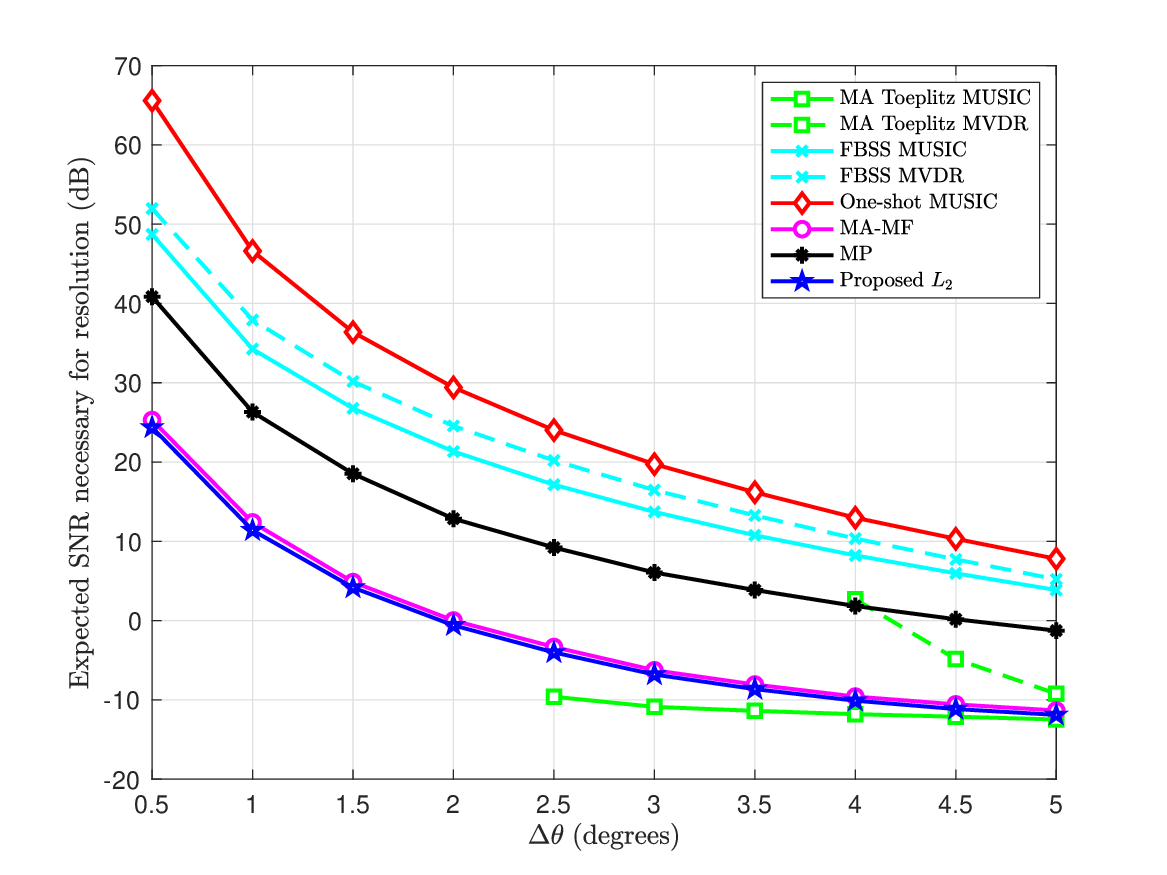}
    } \\
    \subfloat[]{
        \label{fig:l2_M_32}
        \includegraphics[width=0.48\textwidth]{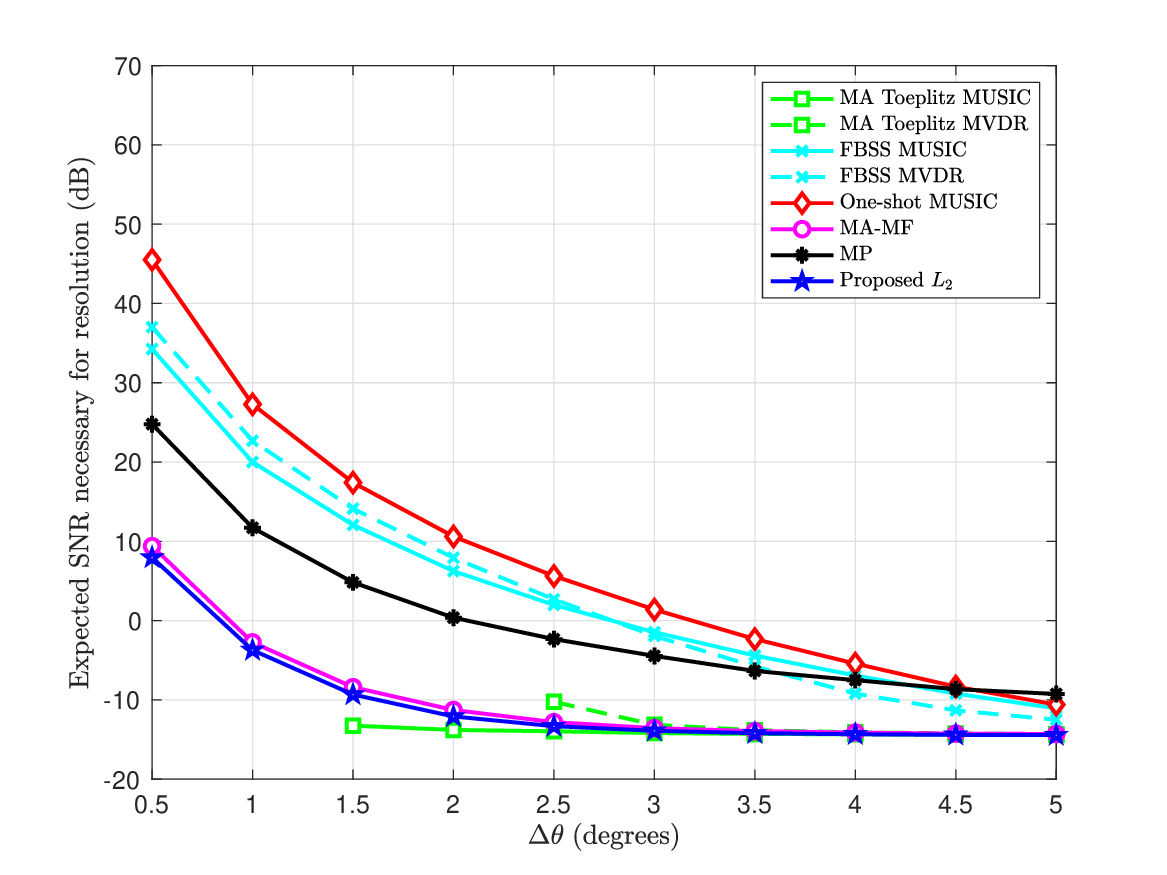}
    }
    \caption{Expected SNR required for successful resolution as a function of source angular separation $\Delta\theta$: (a) $M=16$ and (b) $M=32$.}
    \label{fig:l2_exp_snr_vs_delta_theta}
\end{figure}

Fig. \ref{fig:l2_exp_snr_vs_delta_theta} illustrates the expected SNR required to successfully resolve two sources as a function of their angular separation $\Delta\theta$. In the challenging regime of  small separations $\Delta\theta = 1^\circ$ and $\Delta\theta = 0.5^\circ$, the proposed estimator consistently achieves the lowest required SNR among all considered approaches, followed  closely by MA-MF. Interestingly, multiplicity-averaged Toeplitz covariance–based methods exhibit an identifiability limitation that persists regardless of the SNR when the sources are closely spaced. While increasing the number of antenna elements from $M=16$ (Fig. \ref{fig:l2_exp_snr_vs_delta_theta}(a)) to $M=32$ (Fig. \ref{fig:l2_exp_snr_vs_delta_theta}(b)) offers some relief, these methods still fail to operate in small angular separations. Overall, increasing the array size from $M=16$ to $M=32$ significantly enhances the resolution capability of all methods. 
The relative performance ordering remains unchanged, with the proposed estimator maintaining a consistent and growing advantage in the closely spaced source regime.

\begin{figure*}[t]
    \centering
    \begin{minipage}[t]{0.48\textwidth}
        \centering
        \subfloat[]{
            \includegraphics[width=\textwidth]{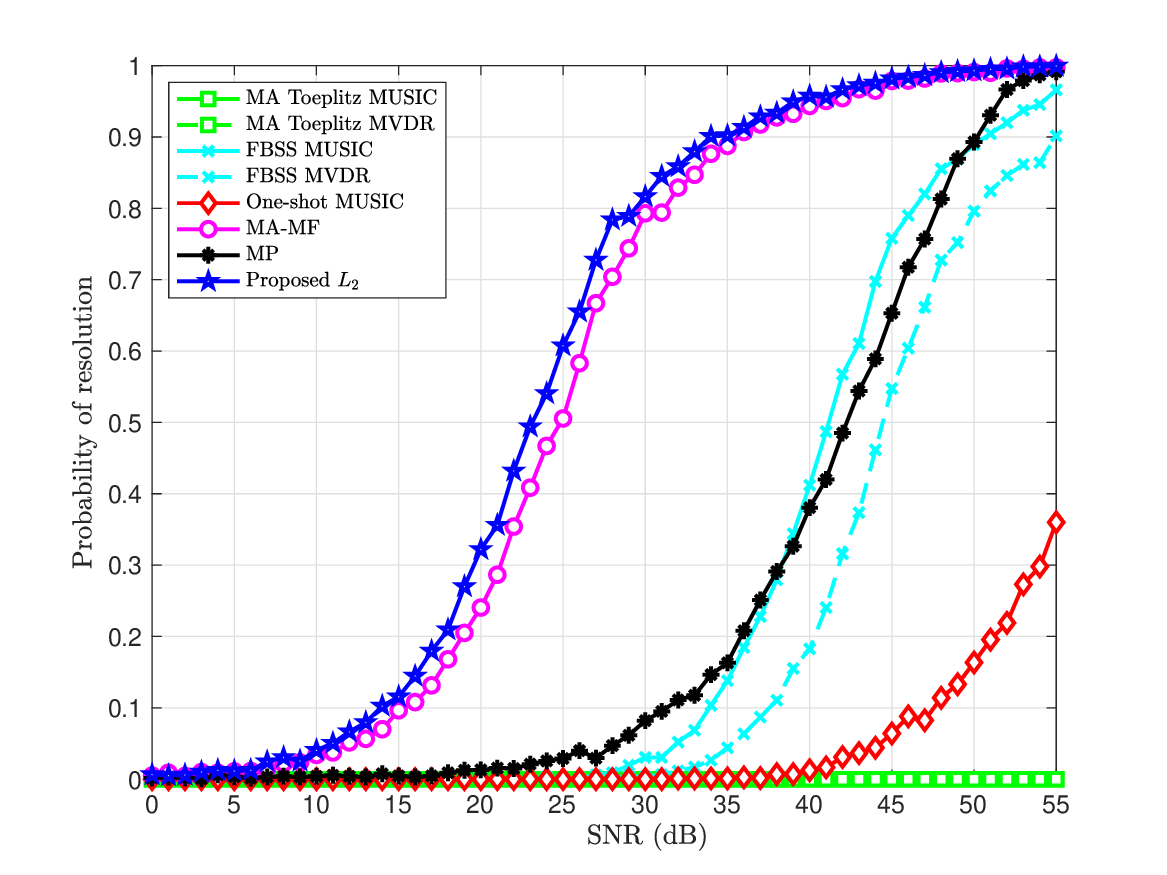}
        } \\
        \subfloat[]{
            \includegraphics[width=\textwidth]{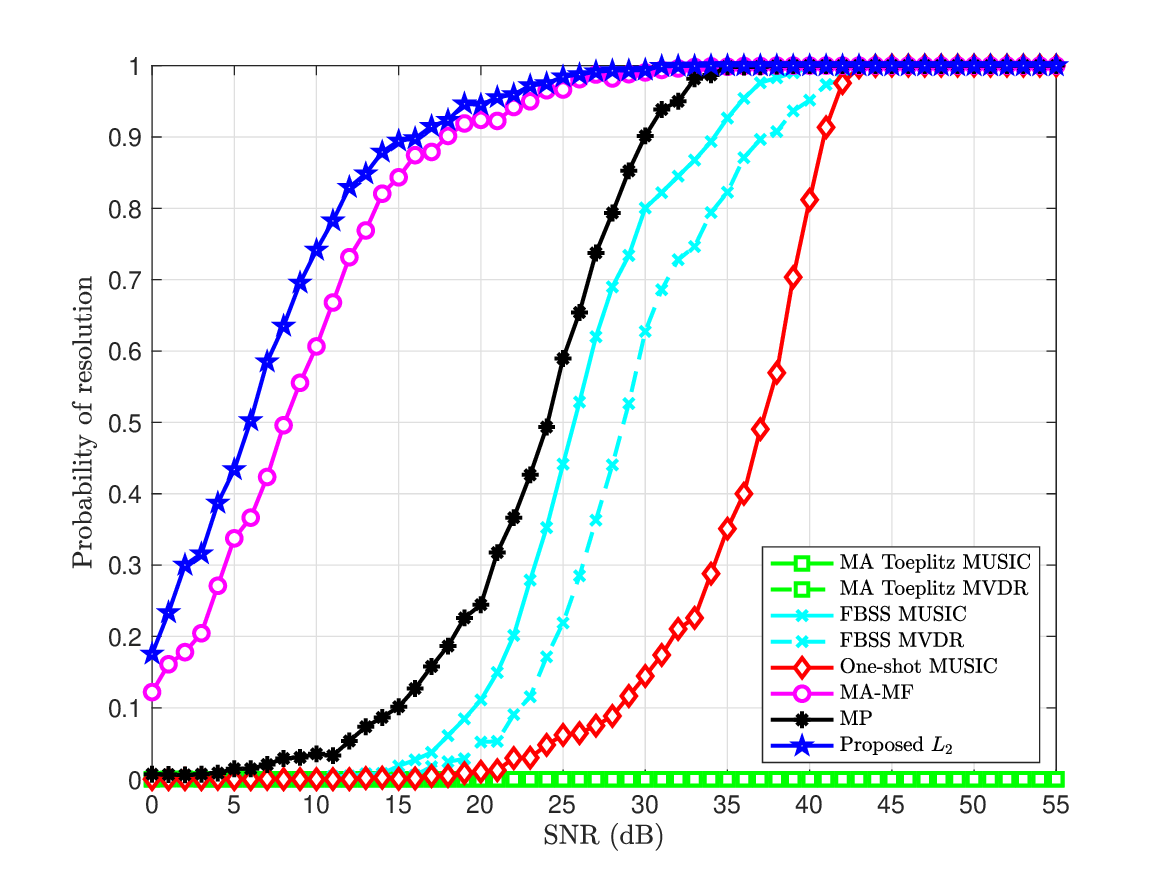}
        }
        \caption{Probability of resolution versus SNR for $\Delta\theta = 1^\circ$: (a) $M=16$, and (b) $M=32$.}
        \label{fig:l2_prob_res_vs_snr_dtheta_1}
    \end{minipage}
    \hfill
    \begin{minipage}[t]{0.48\textwidth}
        \centering
        \subfloat[]{
            \includegraphics[width=\textwidth]{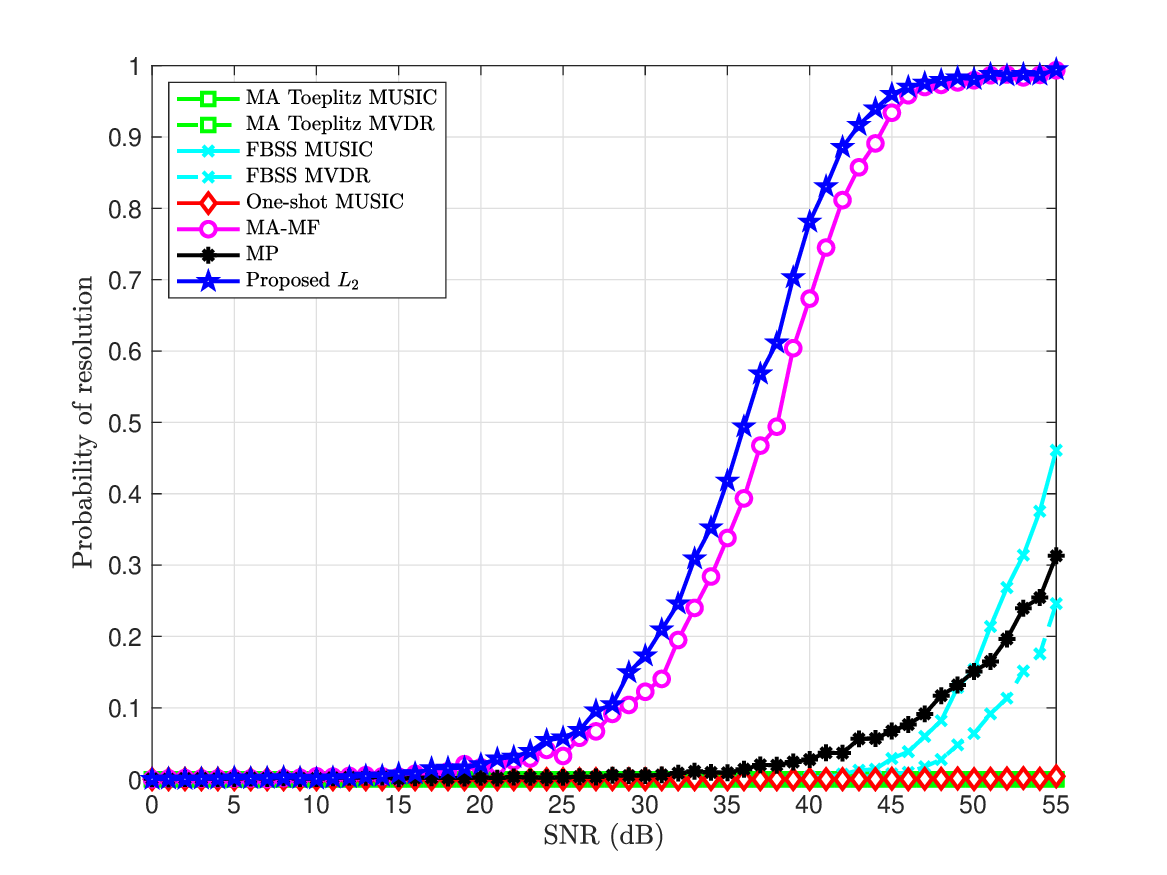}
        } \\
        \subfloat[]{
            \includegraphics[width=\textwidth]{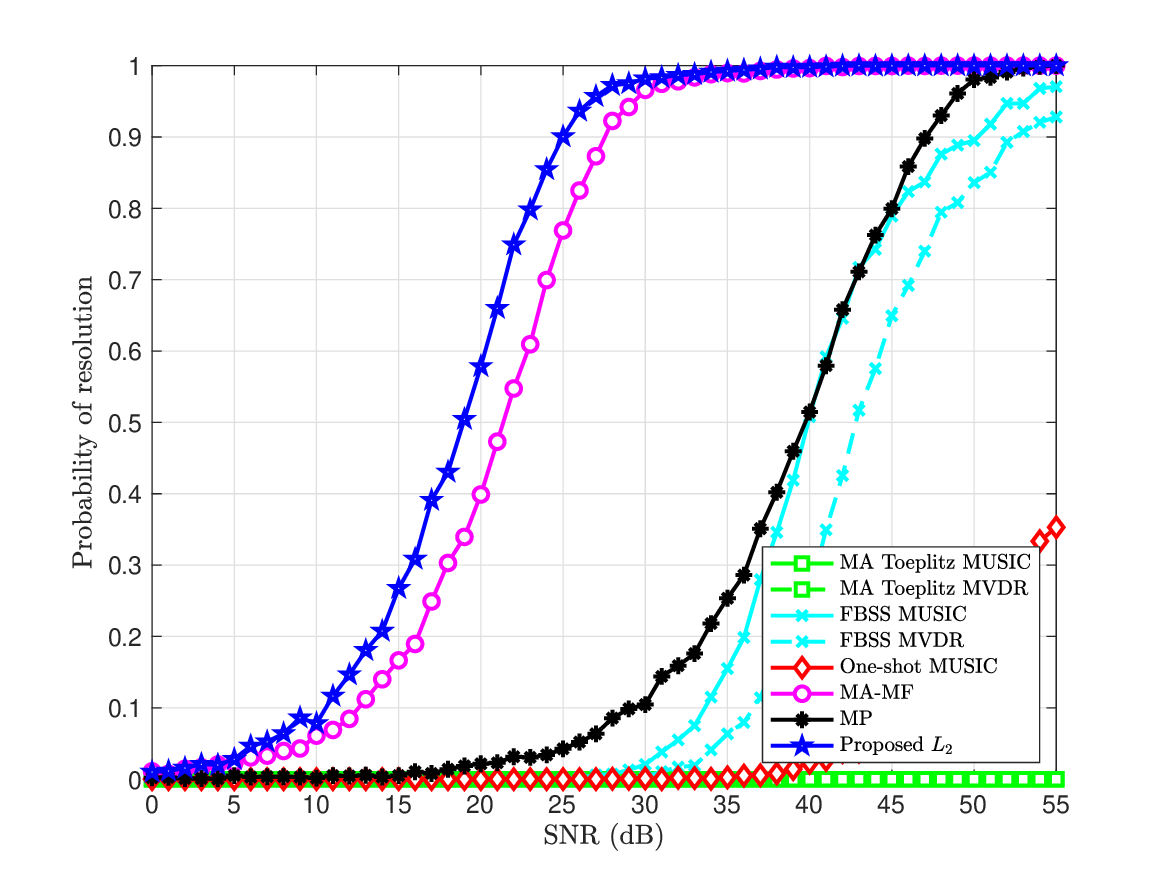}
        }
        \caption{Probability of resolution versus SNR for $\Delta\theta = 0.5^\circ$: (a) $M=16$, and (b) $M=32$.}
        \label{fig:l2_prob_res_vs_snr_dtheta_0.5}
    \end{minipage}
\end{figure*}

Next, we examine the probability of successful resolution as a function of SNR. Fig. \ref{fig:l2_prob_res_vs_snr_dtheta_1} plots the resolution probability versus SNR for fixed angular separation $\Delta\theta= 1^\circ$, for $M=16$ (Fig. \ref{fig:l2_prob_res_vs_snr_dtheta_1}(a)) and $M=32$ (Fig. \ref{fig:l2_prob_res_vs_snr_dtheta_1}(b)) sensing elements.
The proposed estimator under the $L_2$-norm formulation consistently achieves the highest probability of resolution across the entire SNR range. For $M=16$, a notable operating point occurs at approximately $25$dB where the proposed estimator attains about $10\%$ higher resolution probability compared to the closest competing method, MA-MF. 
For $M=32$, at approximately$10$ dB the proposed estimator attains about $15\%$ higher resolution probability than the MA-MF method. At $15$dB, the estimator reaches nearly $90\%$ probability of resolution.
In Fig. \ref{fig:l2_prob_res_vs_snr_dtheta_0.5}, we repeat the same study for $\Delta\theta = 0.5^\circ$  representing the highly challenging scenario of two sources only half-a-degree apart. The same general conclusions can be drawn, while the performance gap of the proposed estimator increases. Remarkably, at $20$dB the $L_2$-norm Hankel-decomposition estimator is nearly $20\%$ more likely to resolve than MA-MF. The estimator reaches $90\%$ probability of resolving the two sources at $25$dB.
%

\begin{figure*}[t]
    \centering
    \begin{minipage}[t]{0.48\textwidth}
        \centering
        \subfloat[]{
            \includegraphics[width=\textwidth]{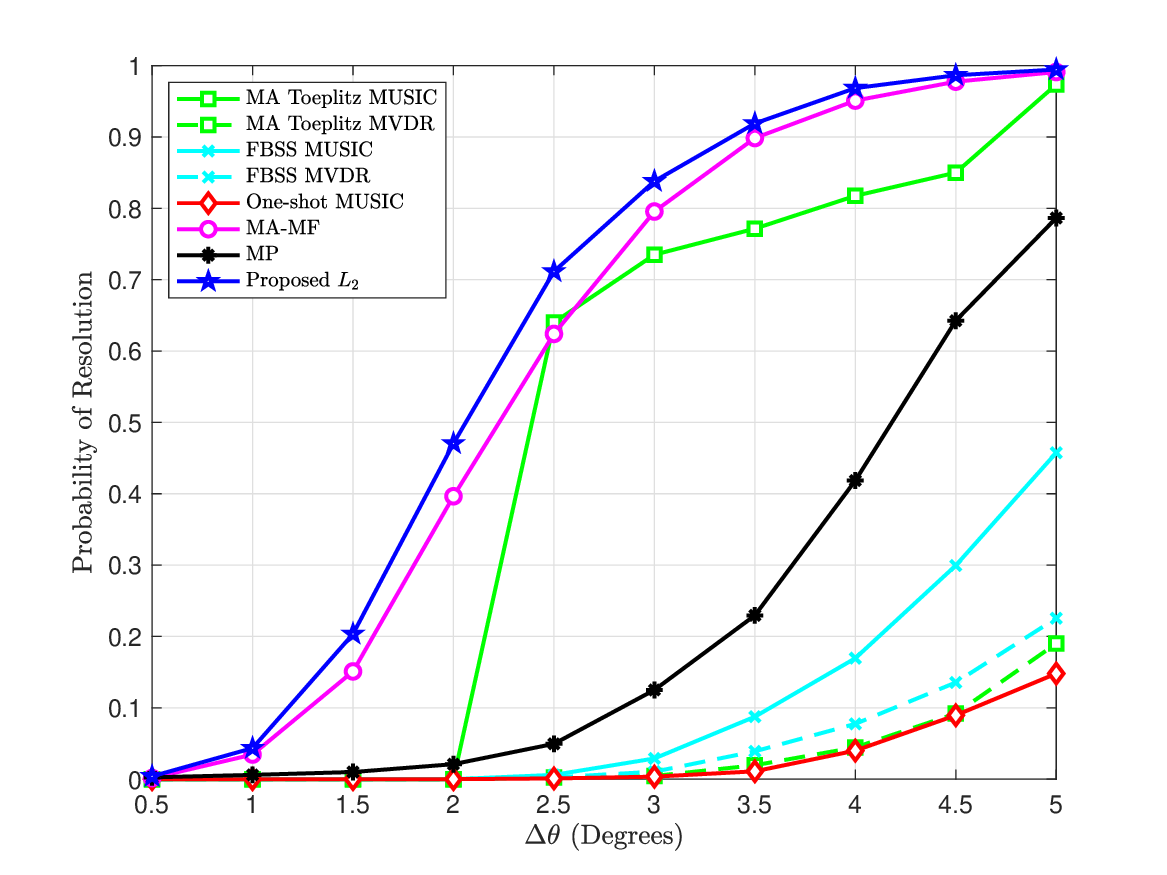}
        } \\
        \subfloat[]{
            \includegraphics[width=\textwidth]{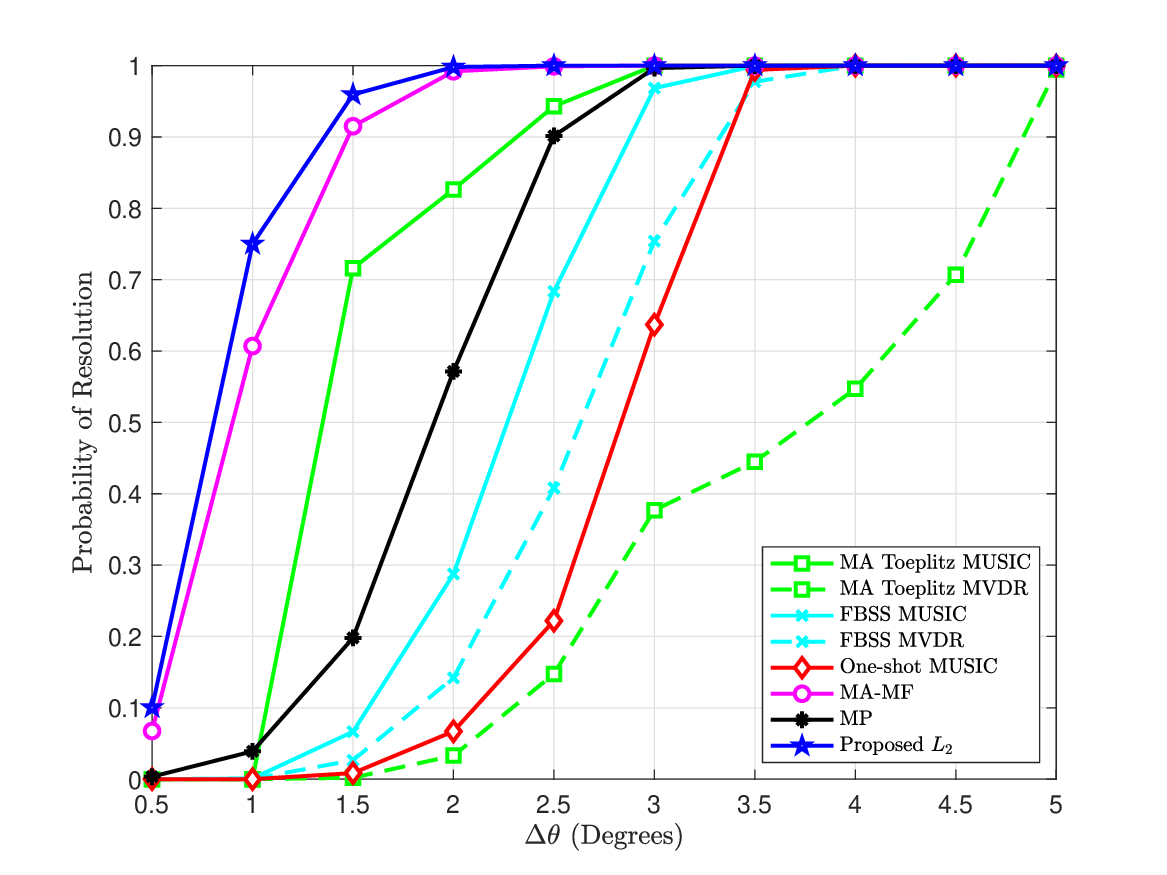}
        }
        \caption{Probability of resolution versus $\Delta\theta$ for $\text{SNR}=10$ dB: (a) $M=16$, and (b) $M=32$.}
        \label{fig:l2_prob_res_vs_dtheta_snr10}
    \end{minipage}
    \hfill
    \begin{minipage}[t]{0.48\textwidth}
        \centering
        \subfloat[]{
            \includegraphics[width=\textwidth]{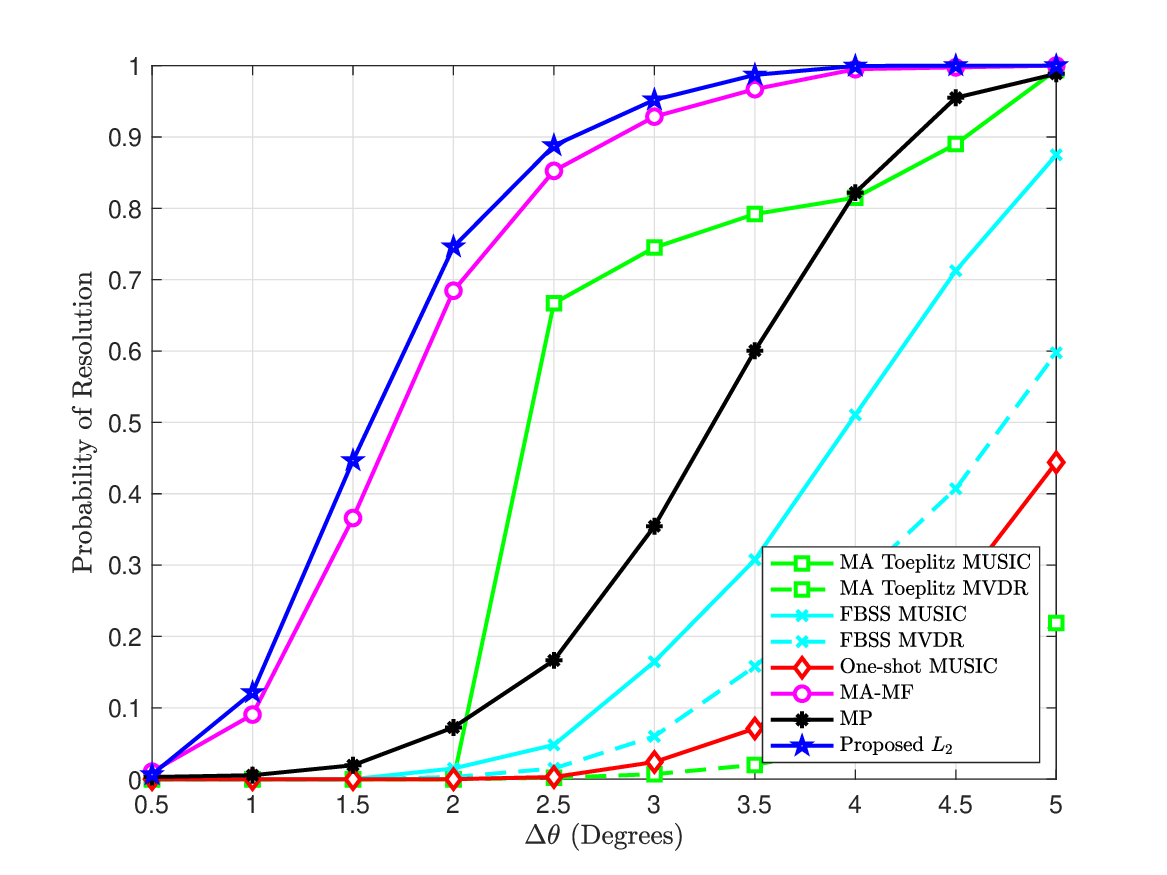}
        } \\
        \subfloat[]{
            \includegraphics[width=\textwidth]{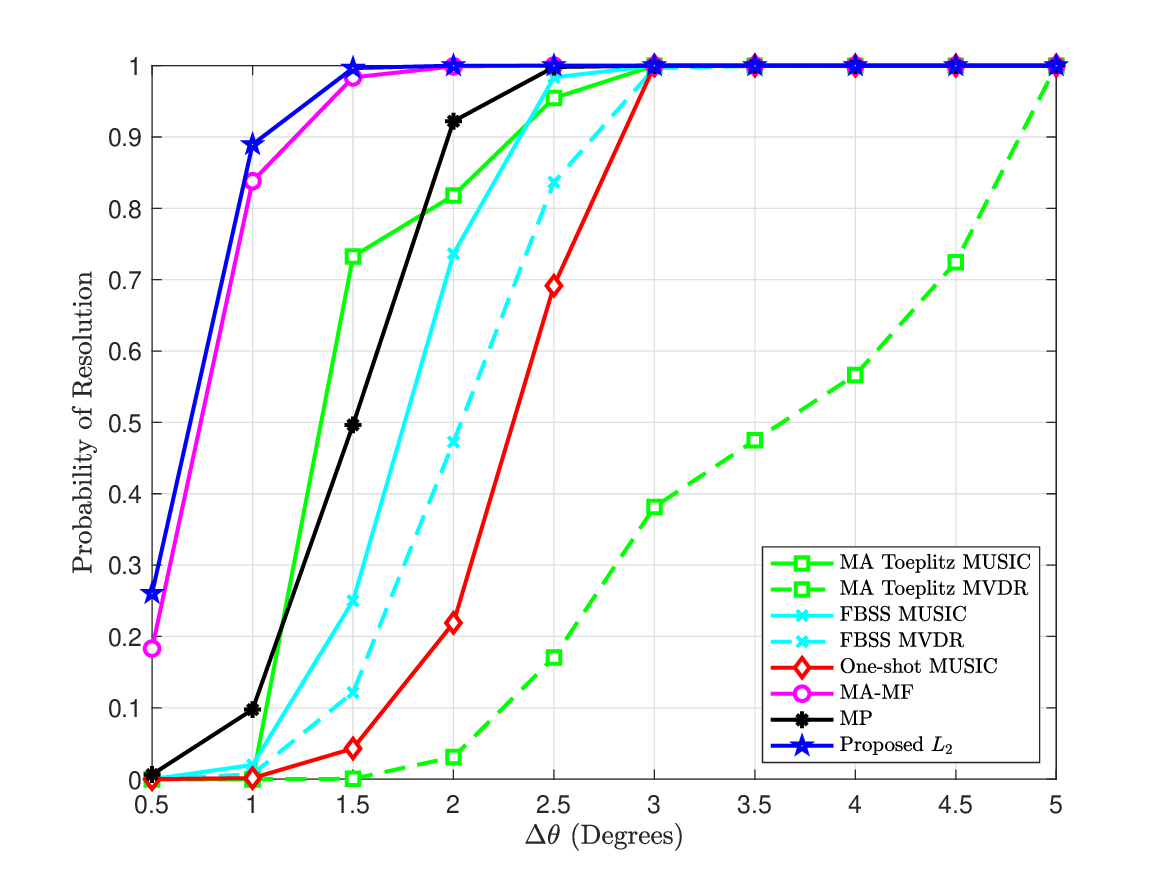}
        }
        \caption{Probability of resolution versus $\Delta\theta$ for $\text{SNR}=15$ dB: (a) $M=16$, and (b) $M=32$.}
        \label{fig:l2_prob_res_vs_dtheta_snr15}
    \end{minipage}
\end{figure*}

Finally, we evaluate the probability of resolution as a function of the angular separation $\Delta\theta$ for fixed SNR levels. Figs. \ref{fig:l2_prob_res_vs_dtheta_snr10} and \ref{fig:l2_prob_res_vs_dtheta_snr15} present the resolution probability versus angular separation for $\text{SNR}=10$dB and $\text{SNR}=15$dB, respectively. In both scenarios, the proposed DoA estimator consistently achieves higher probability of resolution across all values of $\Delta\theta$, underscoring its superior resolution capability compared to competing methods. 

In addition to the $L_2$-norm formulation, we perform the same set of experiments for the $L_1$-norm estimator (Fig. \ref{l1_doa_alg}) to assess its robustness under adverse sensing conditions, particularly in the presence of impulsive noise. We consider a challenging noise environment modeled via a Bernoulli–Gaussian contamination process, which captures sporadic high-power interference. With probability $p>0$, each sensor measurement is affected independently by high-power complex Gaussian noise $n  \sim \mathcal{CN}(0,\sigma_2^2=200)$. With probability $1-p$, each sensor measurement is affected independently by benign lower-power complex Gaussian noise $n  \sim \mathcal{CN}(0,\sigma_1^2=1)$. In this context, 
the per-antenna-element SNR is given by
$\mathrm{SNR} 
= \frac{|x|^2}{(1-p)\sigma_1^2+p\sigma_2^2}$
and the SNR value is controlled by setting $x = \sqrt{10^{SNR / 10}\left[(1-p) \sigma_1^2+p \sigma_2^2\right]} e^{j \phi}$ where $\phi$ is drawn uniformly from $[0, 2\pi]$.

\begin{figure*}[t]
    \centering
    \begin{minipage}[t]{0.48\textwidth}
        \centering
        \subfloat[]{
            \includegraphics[width=\textwidth]{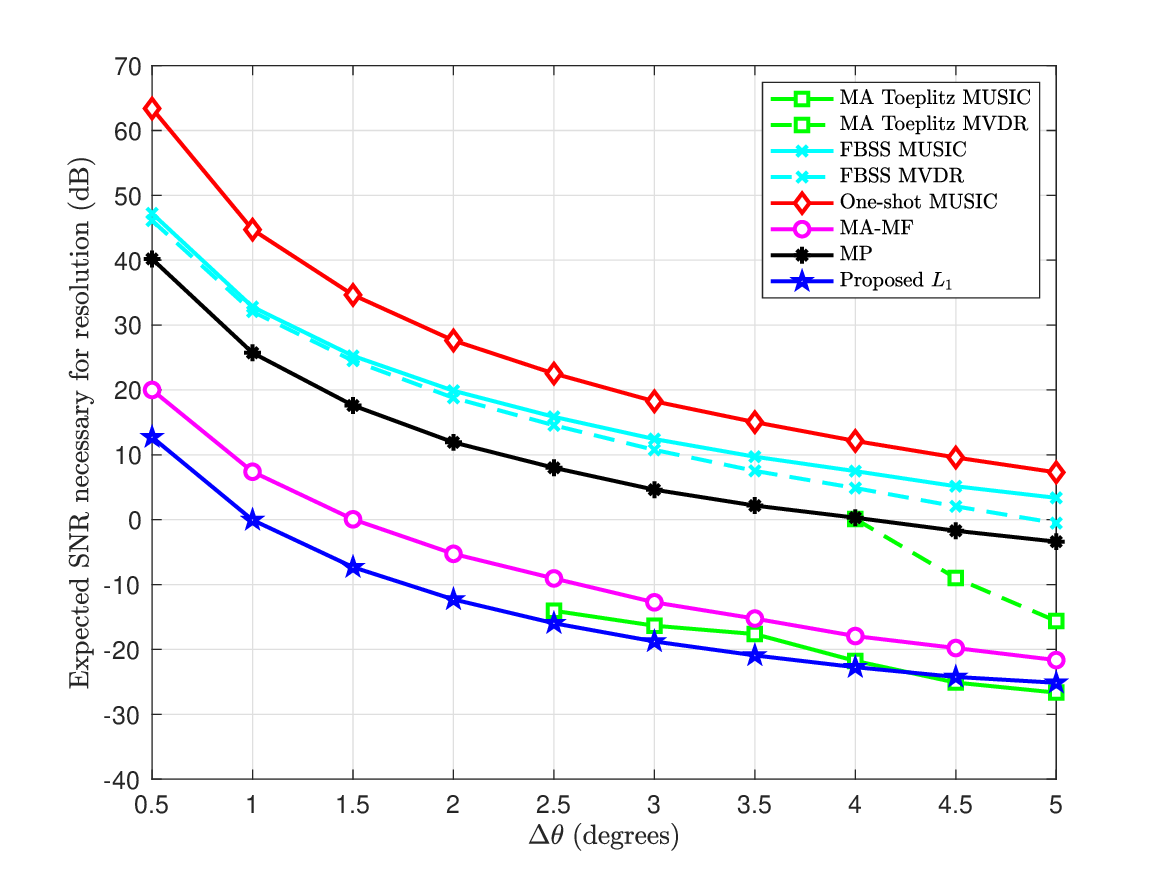}
        } \\
        \subfloat[]{
            \includegraphics[width=\textwidth]{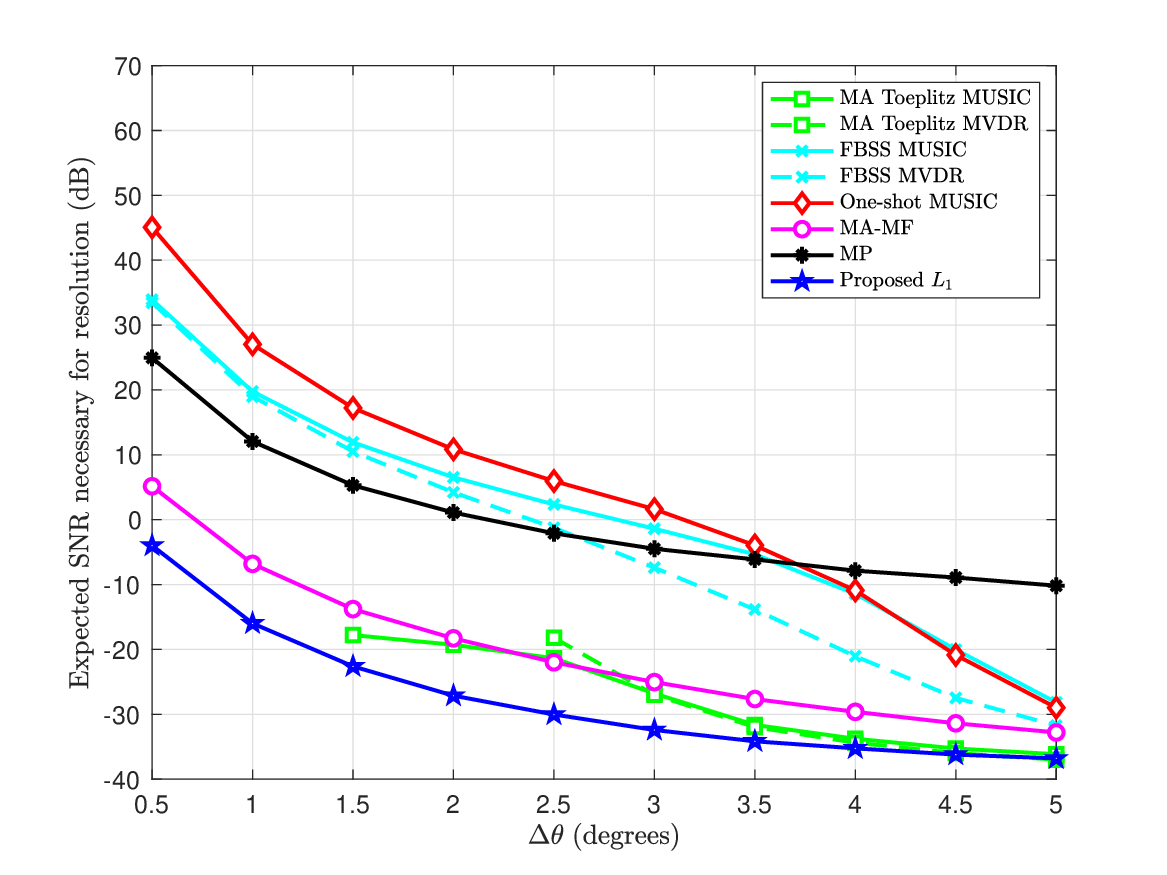}
        }
        \caption{Expected SNR required for successful resolution as a function of source angular separation $\Delta\theta$, impulse probability $p=0.1$: (a) $M=16$, and (b) $M=32$.}
        \label{fig:l1_exp_snr_vs_delta_theta_p0.1}
    \end{minipage}
    \hfill
    \begin{minipage}[t]{0.48\textwidth}
        \centering
        \subfloat[]{
            \includegraphics[width=\textwidth]{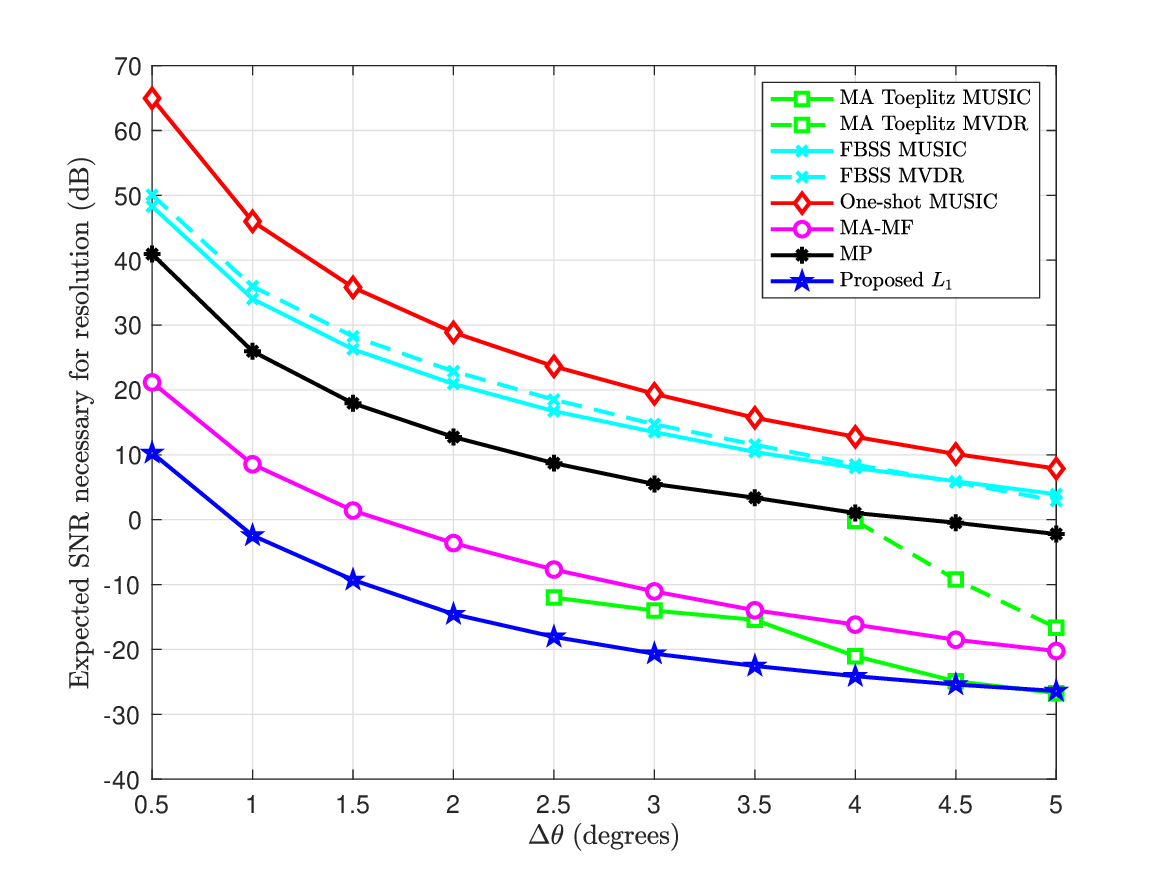}
        } \\
        \subfloat[]{
            \includegraphics[width=\textwidth]{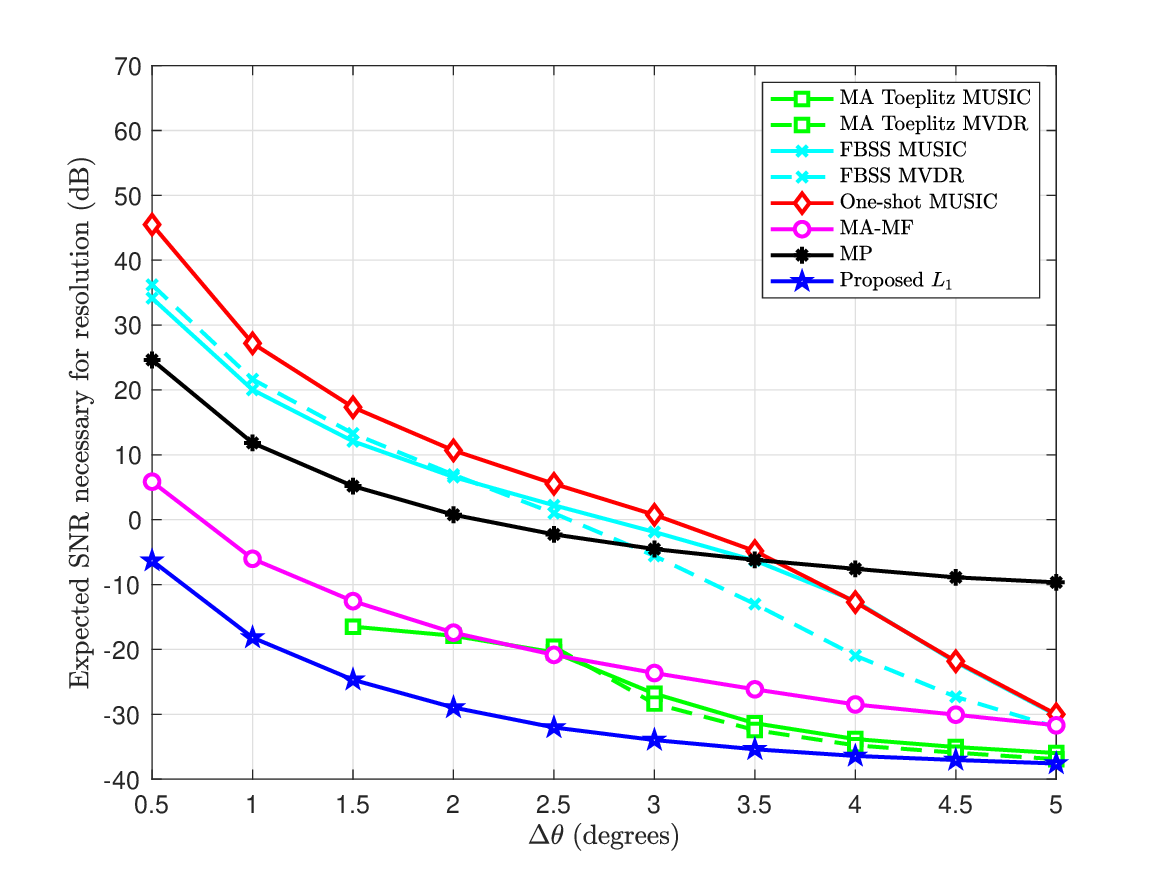}
        }
        \caption{Expected SNR required for successful resolution as a function of source angular separation $\Delta\theta$, impulse probability $p=0.25$: (a) $M=16$, and (b) $M=32$.}
        \label{fig:l1_exp_snr_vs_delta_theta_p0.25}
    \end{minipage}
\end{figure*}

Figs. \ref{fig:l1_exp_snr_vs_delta_theta_p0.1} and \ref{fig:l1_exp_snr_vs_delta_theta_p0.25} show the expected SNR required for successful resolution as a function of angular separation $\Delta\theta$ under impulsive noise with contamination probabilities $p=0.1$ and $p=0.25$, respectively. The proposed $L_1$-norm based estimator consistently achieves the lowest required SNR, with the most pronounced gains observed in the super-resolution regime $\Delta\theta < 1.5^\circ$ demonstrating its robustness to impulsive noise sensing environments. As an example, at $\Delta\theta = 1^\circ$ and moderate contamination probability $p=0.1$, the proposed estimator requires approximately $8$dB and $10$dB lower SNR than the closest competing method for $M=16$ and $M=32$ sensors, respectively. As the impulsive noise becomes more severe ($p=0.25$), the gains are even more significant exceeding $10$dB for $M=16$ and $13$dB for $M=32$ sensors. 

\begin{figure*}[t]
    \begin{minipage}[t]{0.48\textwidth}
        \centering
        \subfloat[]{
            \includegraphics[width=\textwidth]{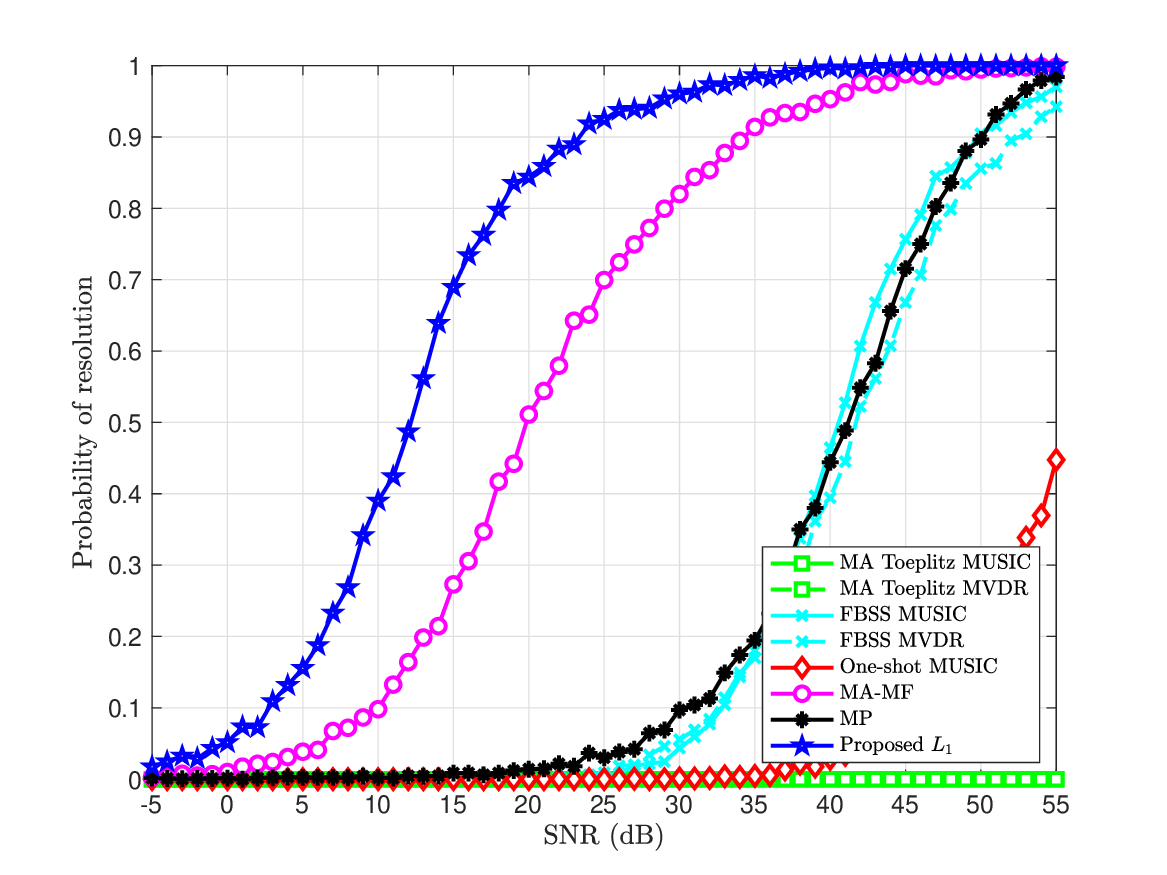}
        } \\
        \subfloat[]{
            \includegraphics[width=\textwidth]{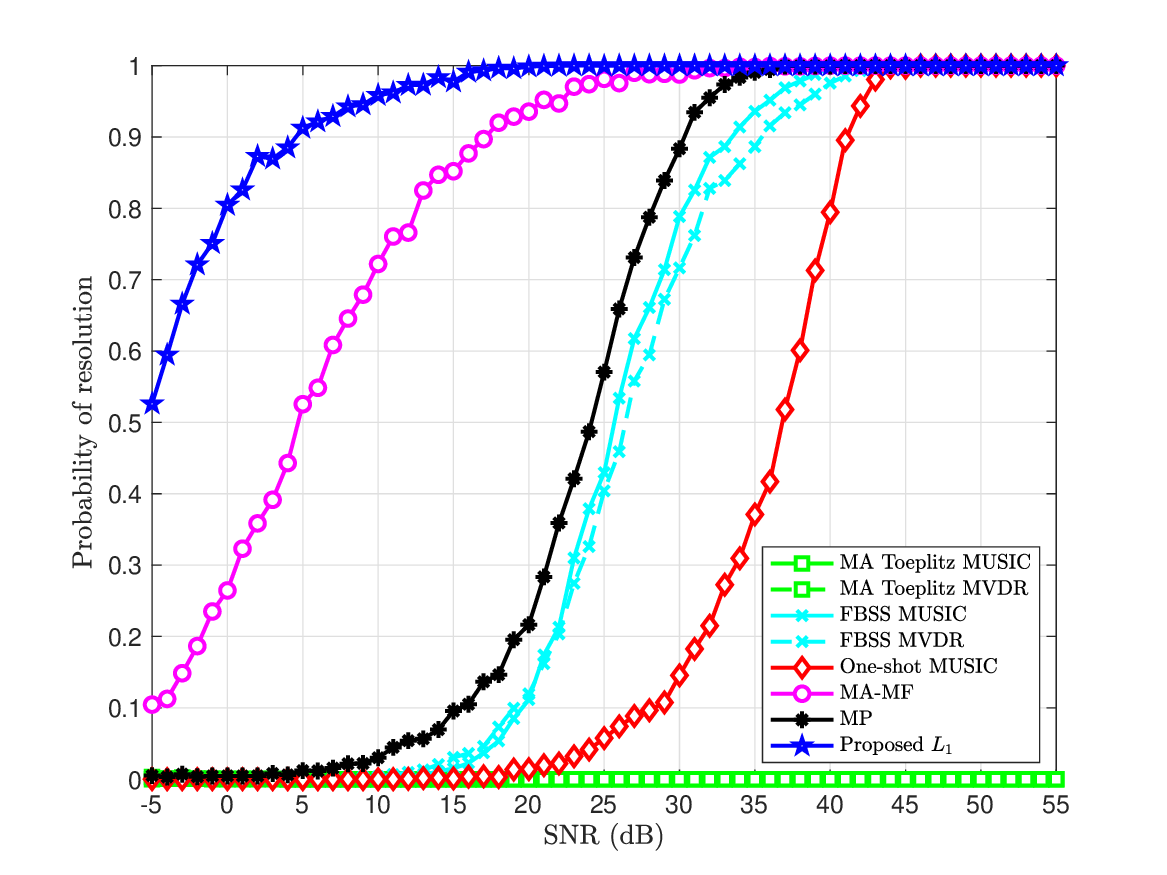}
        }
        \caption{Probability of resolution versus SNR for $\Delta\theta = 1^\circ$, impulse probability $p=0.1$: (a) $M=16$, and (b) $M=32$.}
        \label{fig:l1_prob_res_vs_snr_dtheta_1_p0.1}
    \end{minipage}
    \hfill
    \begin{minipage}[t]{0.48\textwidth}
        \centering
        \subfloat[]{
            \includegraphics[width=\textwidth]{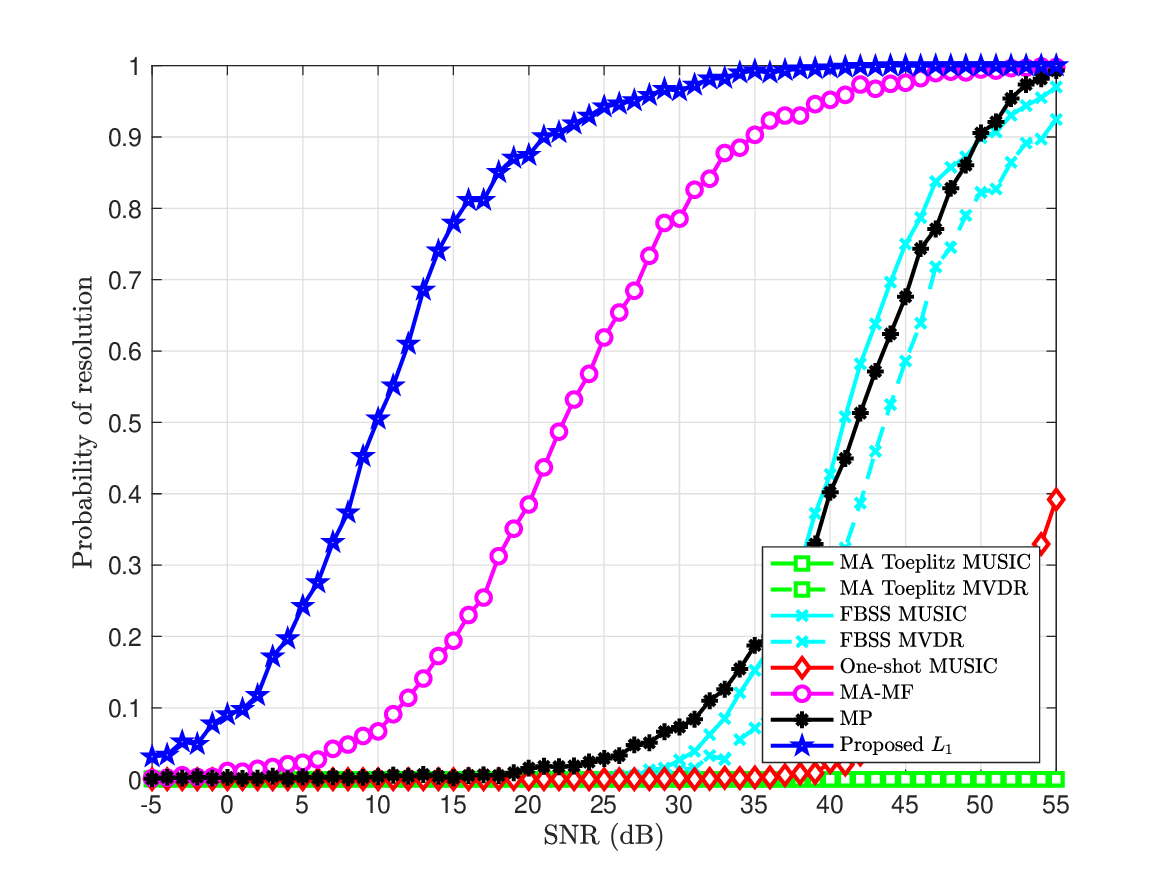}
        } \\
        \subfloat[]{
            \includegraphics[width=\textwidth]{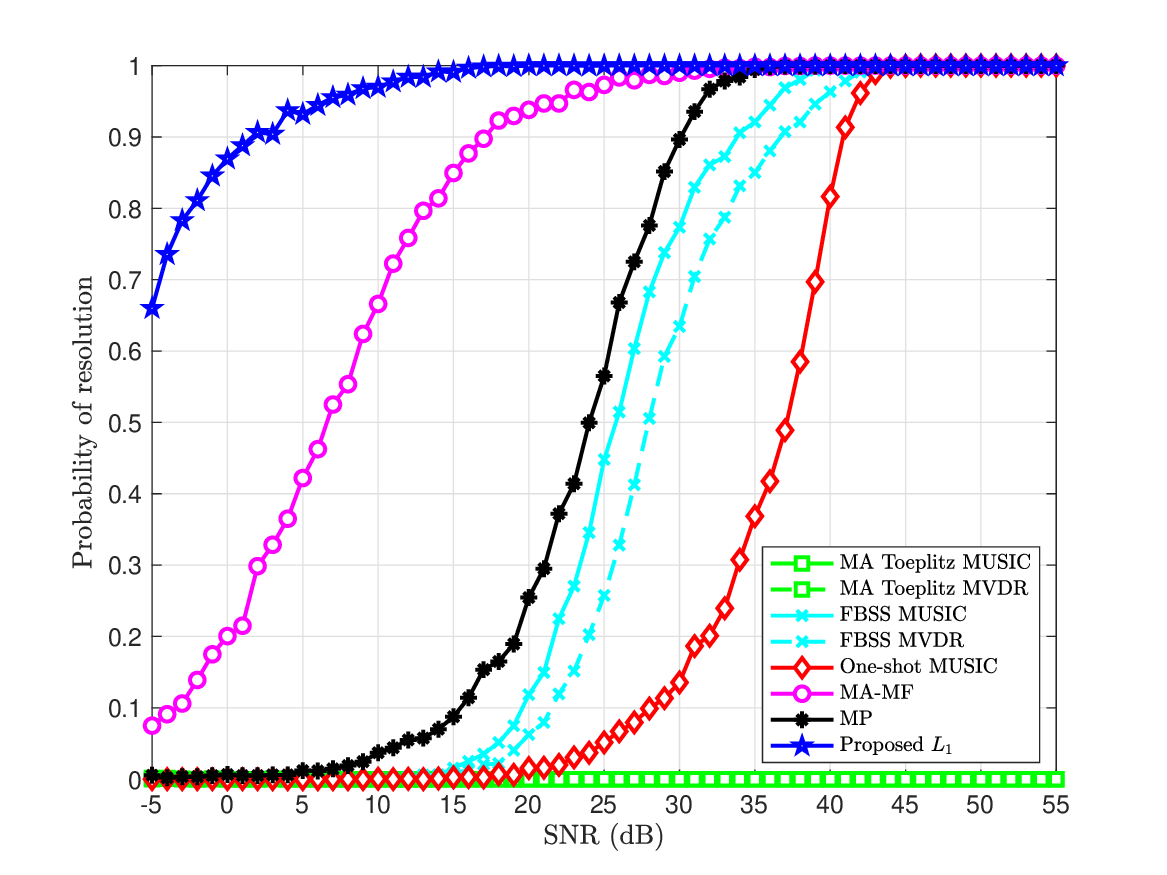}
        }
        \caption{Probability of resolution versus SNR for $\Delta\theta = 1^\circ$, impulse probability $p=0.25$: (a) $M=16$, and (b) $M=32$.}
        \label{fig:l1_prob_res_vs_snr_dtheta_1_p0.25}
    \end{minipage}
\end{figure*}

Figs. \ref{fig:l1_prob_res_vs_snr_dtheta_1_p0.1} and
\ref{fig:l1_prob_res_vs_snr_dtheta_1_p0.25} illustrate the probability of resolution versus SNR for the less demanding angular separation of $\Delta\theta = 1^\circ$ under impulse probabilities $p=0.1$ and $p=0.25$, respectively, for both $M=16$ and $M=32$. As expected, increased angular separation yields an overall improvement in
resolution probability across all methods; nevertheless, the dominance of the
proposed $L_1$-norm-based method is decisive. To illustrate this with a
concrete operating point, consider SNR $= 0$dB with $M = 32$ and $p = 0.25$:
the proposed method already achieves a resolution probability of approximately 
$90\%$, whereas the closest competing approach reaches only approximately
$20\%$, and all remaining methods fail entirely. Taken together, these results
confirm that the proposed method not only outperforms all competitors across
all tested conditions, but does so with an increasingly pronounced margin as
the noise environment becomes more impulsive, underscoring its robustness
and suitability for practical scenarios in which the Gaussian noise assumption
is violated.

\begin{figure*}[t]
    \centering
    \begin{minipage}[t]{0.48\textwidth}
        \centering
        \subfloat[]{
            \includegraphics[width=\textwidth]{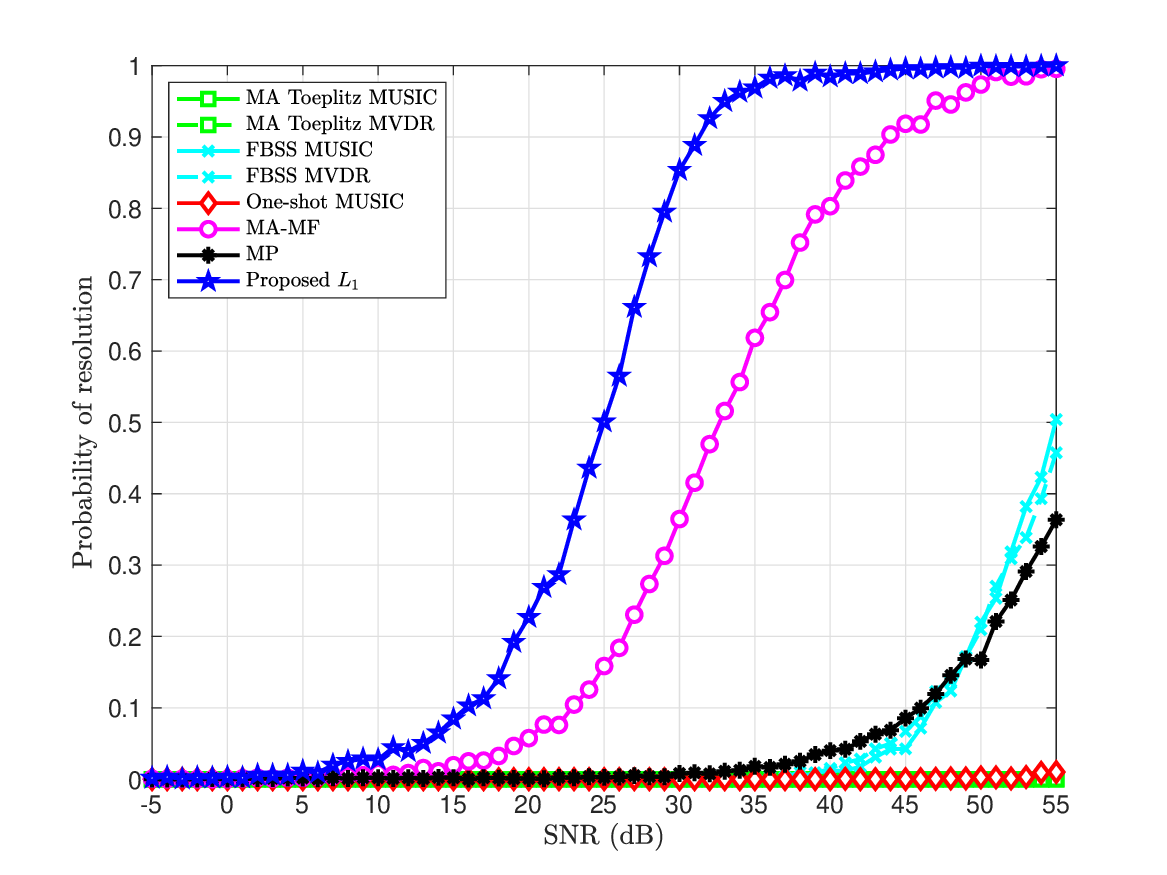}
        } \\
        \subfloat[]{
            \includegraphics[width=\textwidth]{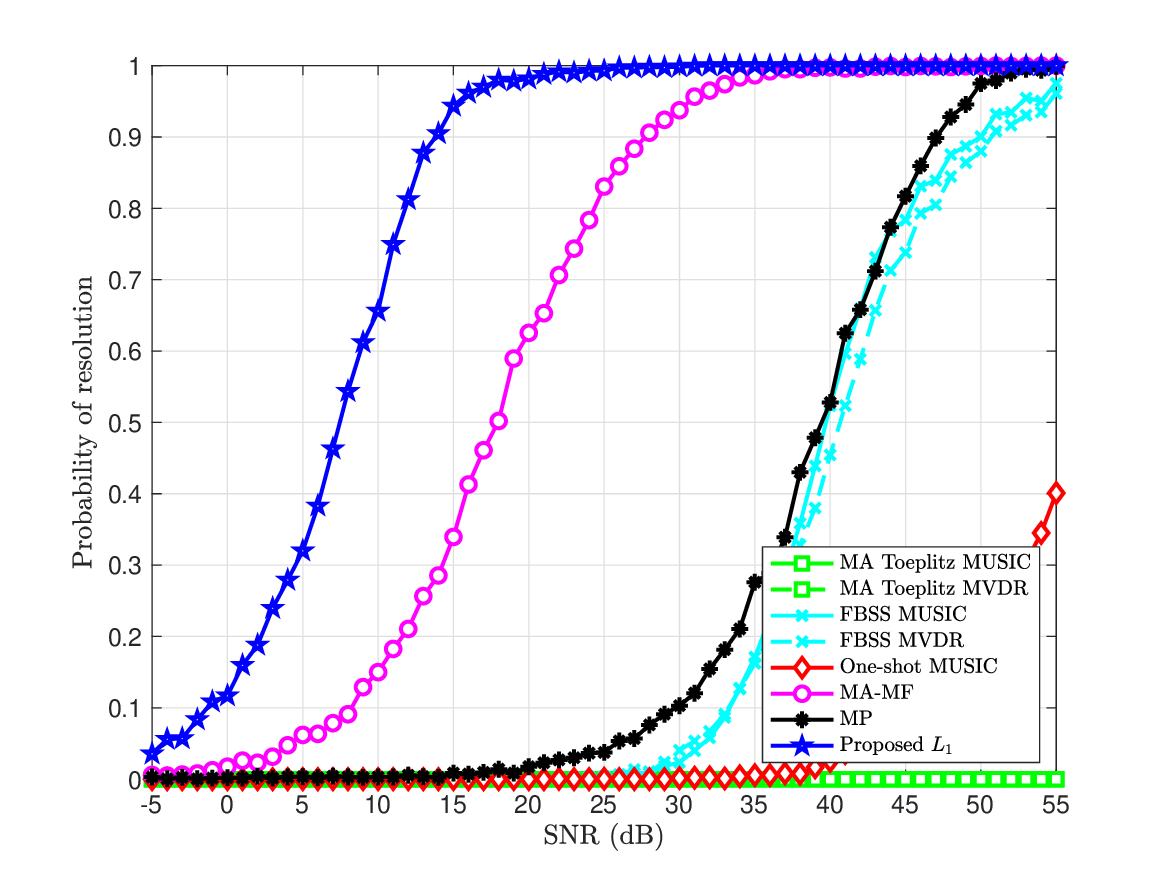}
        }
        \caption{Probability of resolution versus SNR for $\Delta\theta = 0.5^\circ$, impulse probability $p=0.1$: (a) $M=16$, and (b) $M=32$.}
        \label{fig:l1_prob_res_vs_snr_dtheta_0.5_p0.1}
    \end{minipage}
    \hfill
    \centering
    \begin{minipage}[t]{0.48\textwidth}
        \centering
        \subfloat[]{
            \includegraphics[width=\textwidth]{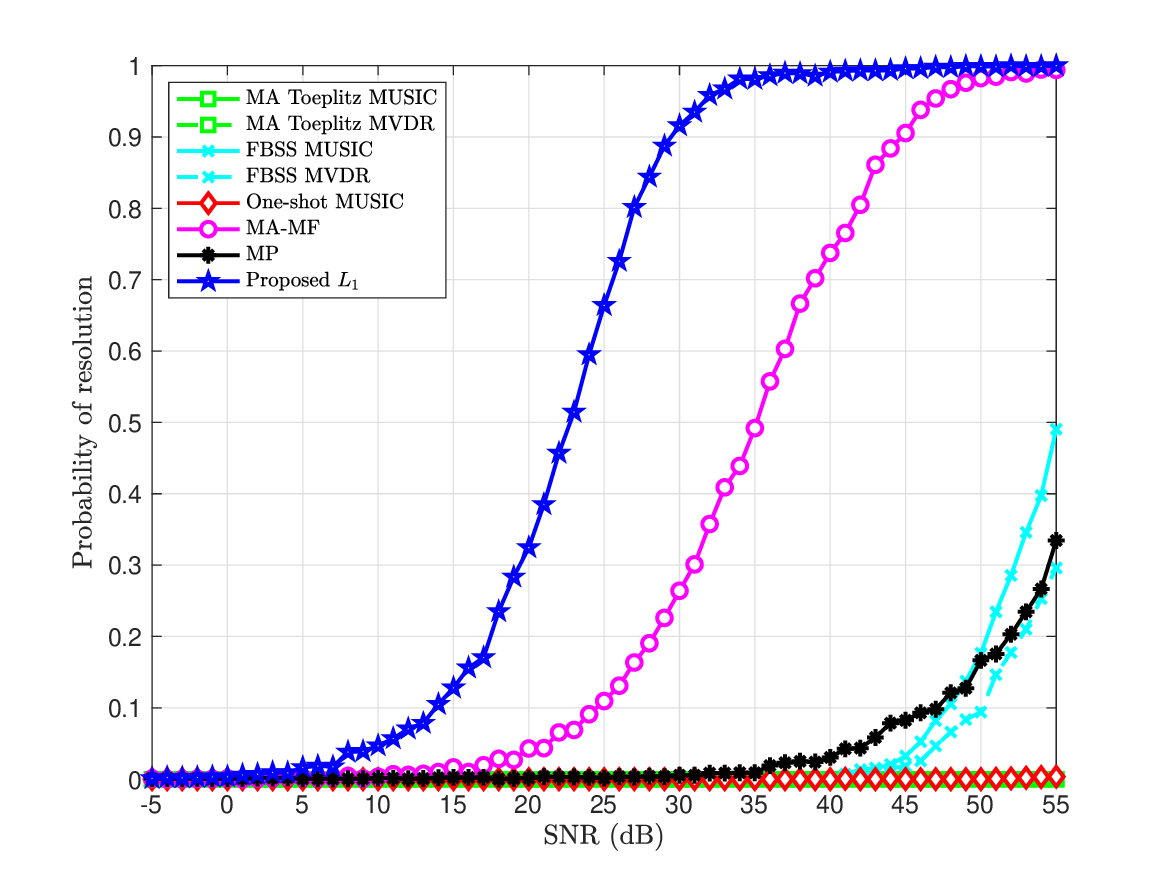}
        } \\
        \subfloat[]{
            \includegraphics[width=\textwidth]{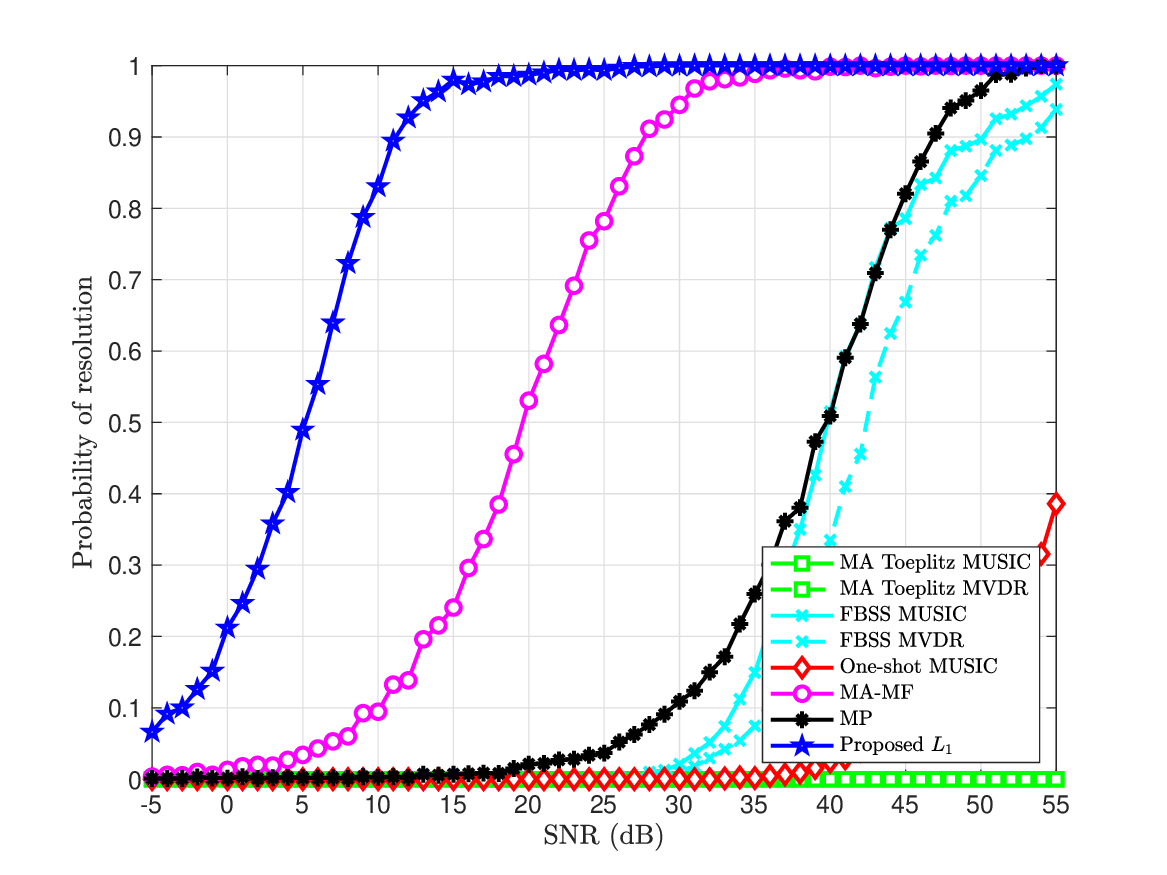}
        }
        \caption{Probability of resolution versus SNR for $\Delta\theta = 0.5^\circ$, impulse probability $p=0.25$: (a) $M=16$, and (b) $M=32$.}
        \label{fig:l1_prob_res_vs_snr_dtheta_0.5_p0.25}
    \end{minipage}
\end{figure*}

We now turn to the more challenging scenario of closely spaced sources with $\Delta\theta = 0.5^\circ$. Figs. \ref{fig:l1_prob_res_vs_snr_dtheta_0.5_p0.1} and
 \ref{fig:l1_prob_res_vs_snr_dtheta_0.5_p0.25} depict the probability of resolution versus SNR under impulse probabilities $p=0.1$ and $p=0.25$, respectively. In all cases, the proposed  $L_1$-norm-based method achieves substantial gains over all competing  frameworks. For $M=16$ at SNR $=30$dB, the proposed method yields 
approximately a $45\%$ improvement in resolution probability over the closest 
competitor when $p=0.1$, which further increases to nearly $60\%$ when $p=0.25$. 
For the larger array $M=32$, the same trends hold but manifest at 
considerably lower SNR values, reflecting the additional spatial diversity 
provided by the larger aperture. Crucially, the performance gap in favor of 
the proposed method widens as $p$ increases from $0.1$ to $0.25$, while the 
competing methods either stagnate or deteriorate. This behavior can be 
explained by two complementary effects. First, under the adopted SNR 
definition with respect to the mixture noise power $(1-p)\sigma_1^2 + 
p\sigma_2^2$, maintaining a fixed SNR at a higher $p$ necessitates a larger 
signal amplitude $|x|$, which inherently favors any robust estimator. Second, 
as $p$ increases the noise distribution becomes increasingly heavy-tailed, 
which is precisely the regime where the $L_1$-norm formulation holds a 
fundamental statistical advantage over methods that rely on second-order 
statistics, whose underlying Gaussian assumption becomes progressively 
violated.

\begin{figure*}[t]
    \centering
    \begin{minipage}[t]{0.48\textwidth}
        \centering
        \subfloat[]{
            \includegraphics[width=\textwidth]{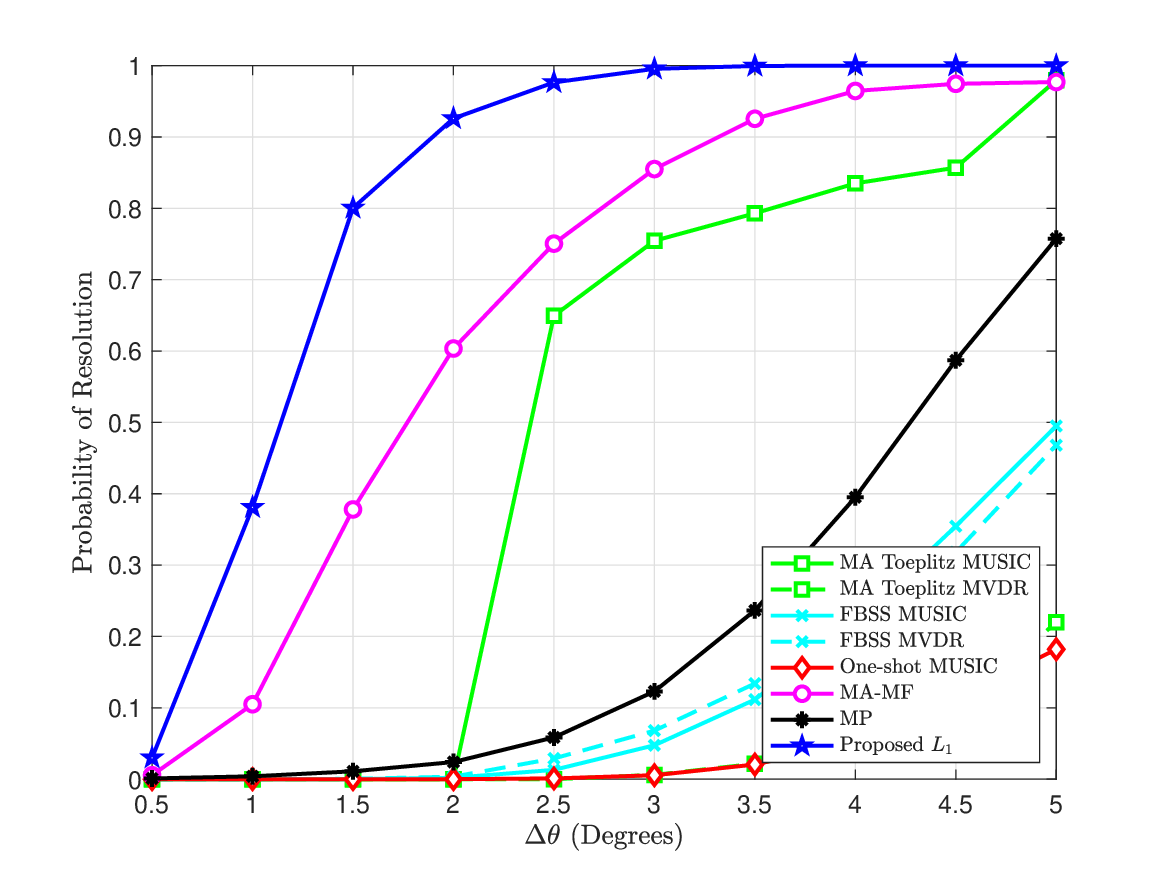}
        } \\
        \subfloat[]{
            \includegraphics[width=\textwidth]{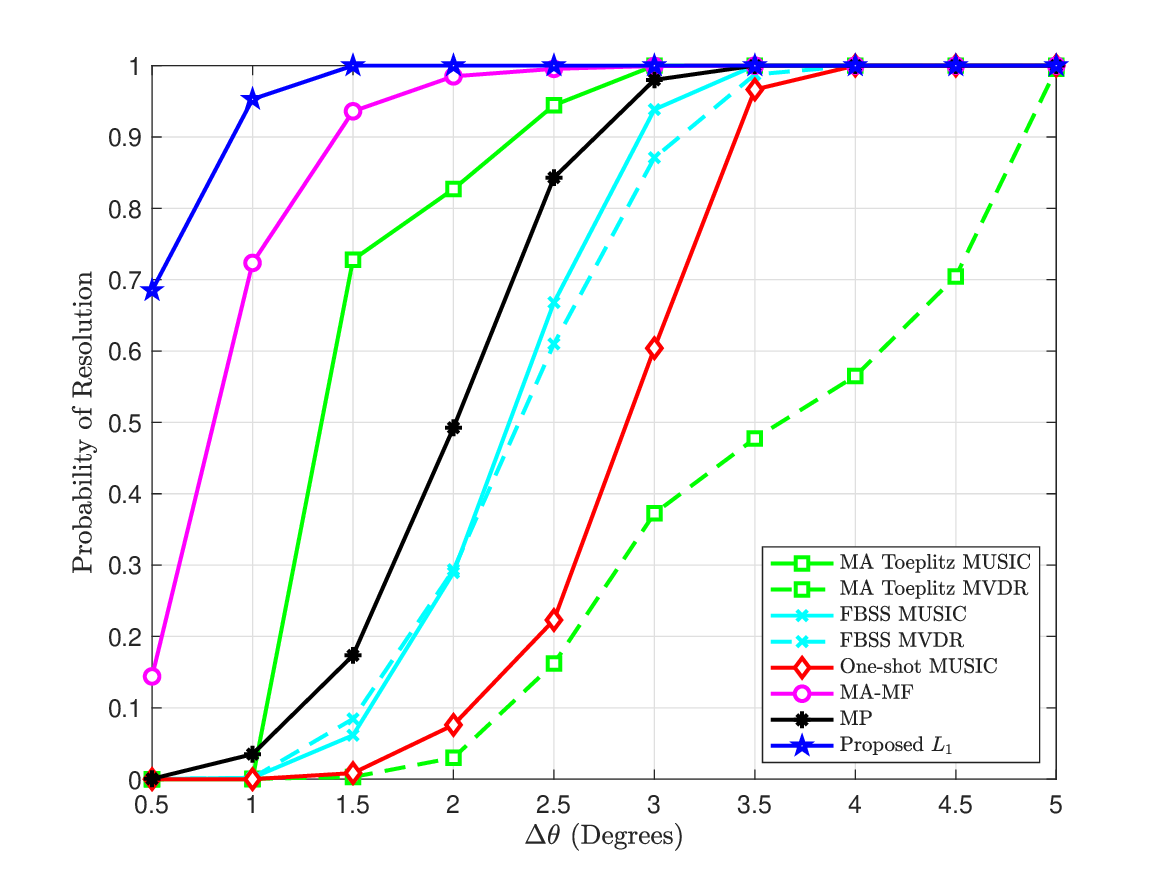}
        }
        \caption{Probability of resolution versus $\Delta\theta$ for $\text{SNR}=10$dB, impulse probability $p=0.1$: (a) $M=16$, and (b) $M=32$.}
        \label{fig:l1_prob_res_vs_dtheta_snr10_p0.1}
    \end{minipage}
    \hfill
    \begin{minipage}[t]{0.48\textwidth}
        \centering
        \subfloat[]{
            \includegraphics[width=\textwidth]{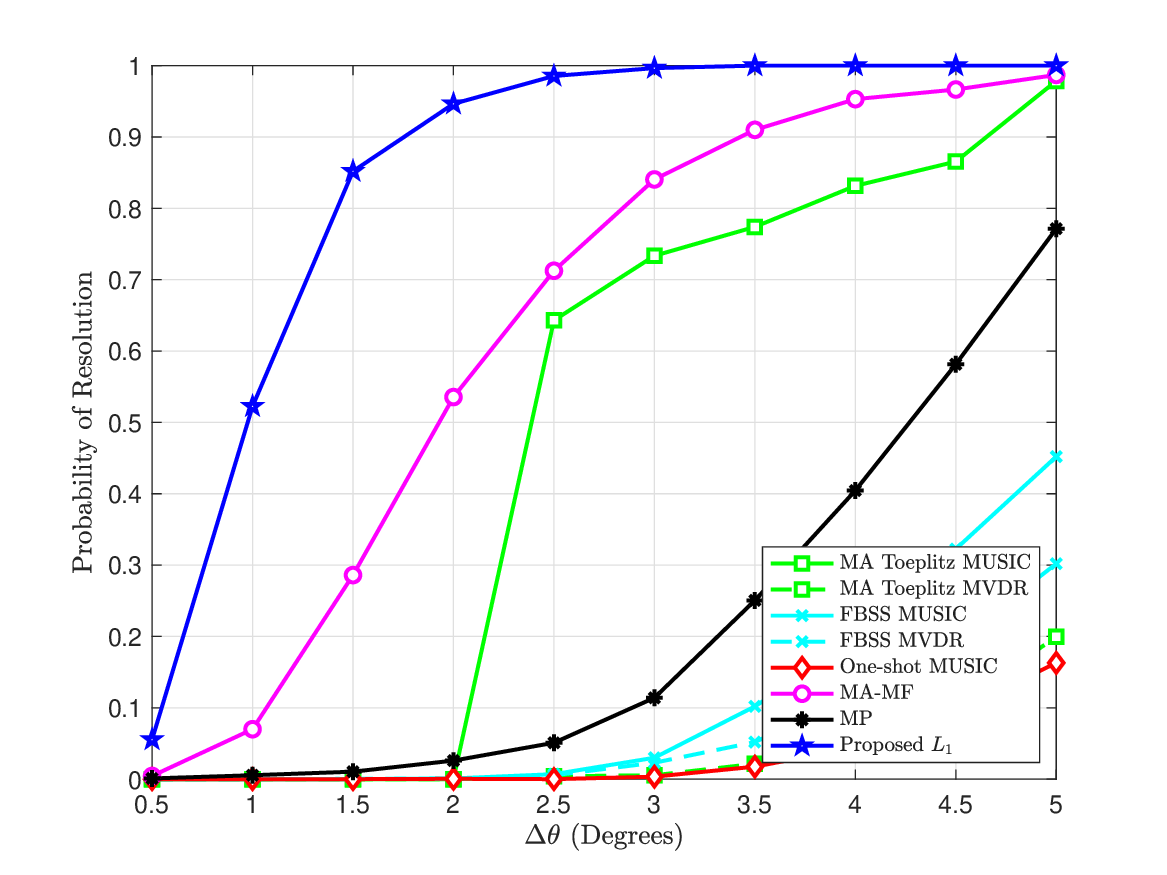}
        } \\
        \subfloat[]{
            \includegraphics[width=\textwidth]{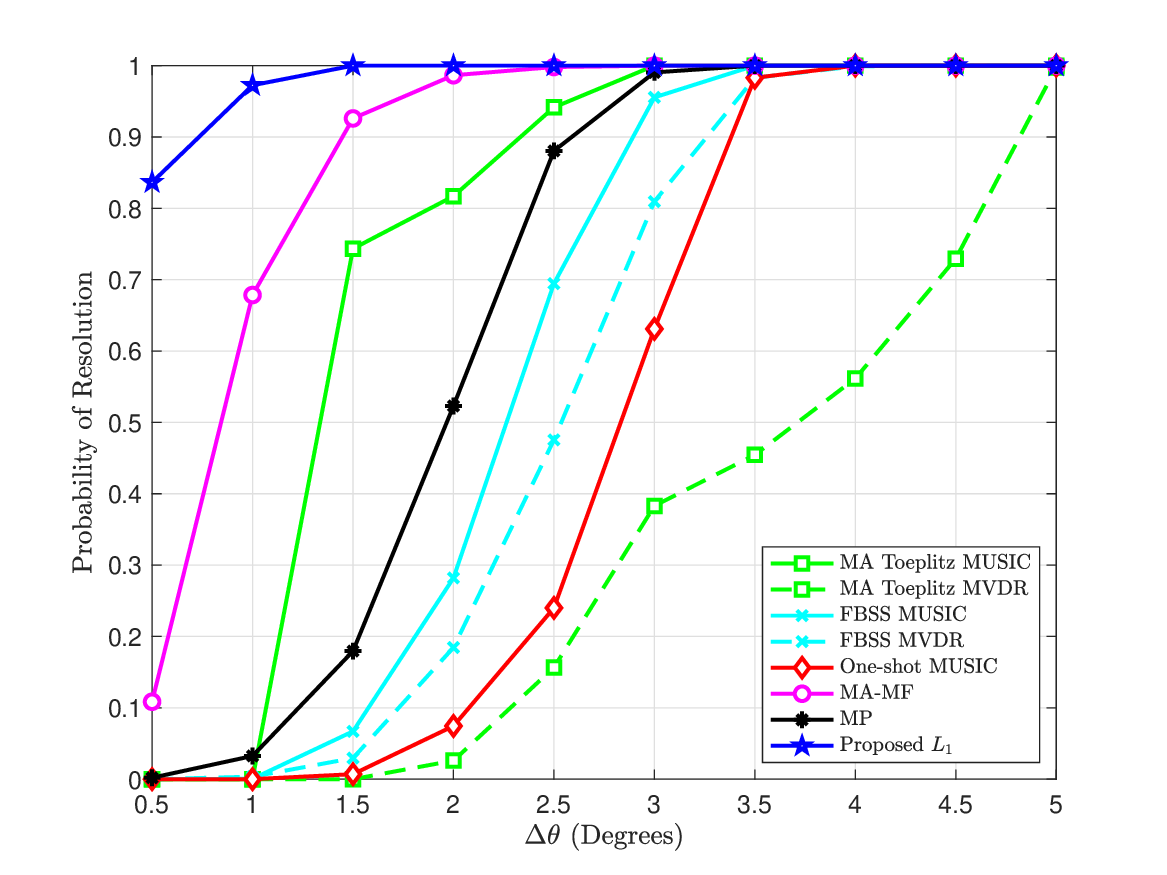}
        }
        \caption{Probability of resolution versus $\Delta\theta$ for $\text{SNR}=10$dB, impulse probability $p=0.25$: (a) $M=16$, and (b) $M=32$.}
        \label{fig:l1_prob_res_vs_dtheta_snr10_p0.25}
    \end{minipage}
\end{figure*}

\begin{figure*}[t]
    \begin{minipage}[t]{0.48\textwidth}
        \centering
        \subfloat[]{
            \includegraphics[width=\textwidth]{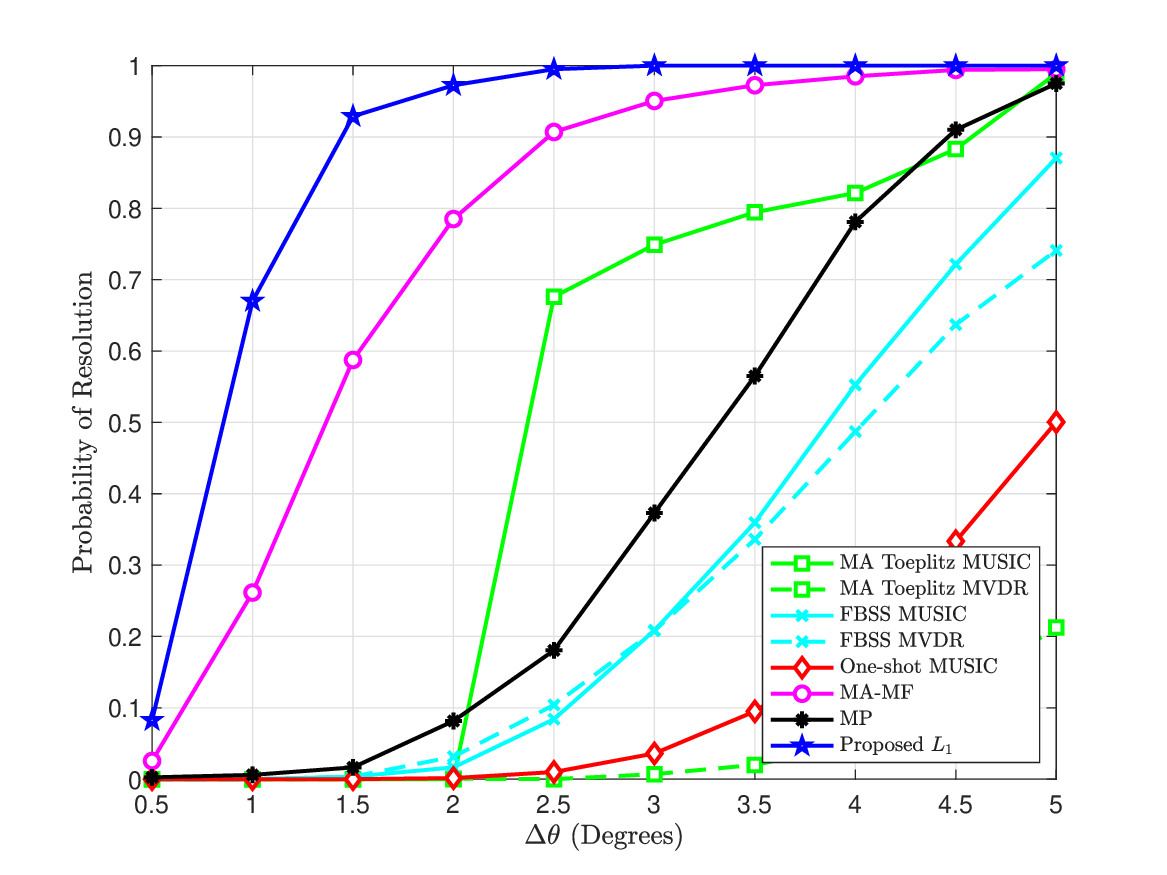}
        } \\
        \subfloat[]{
            \includegraphics[width=\textwidth]{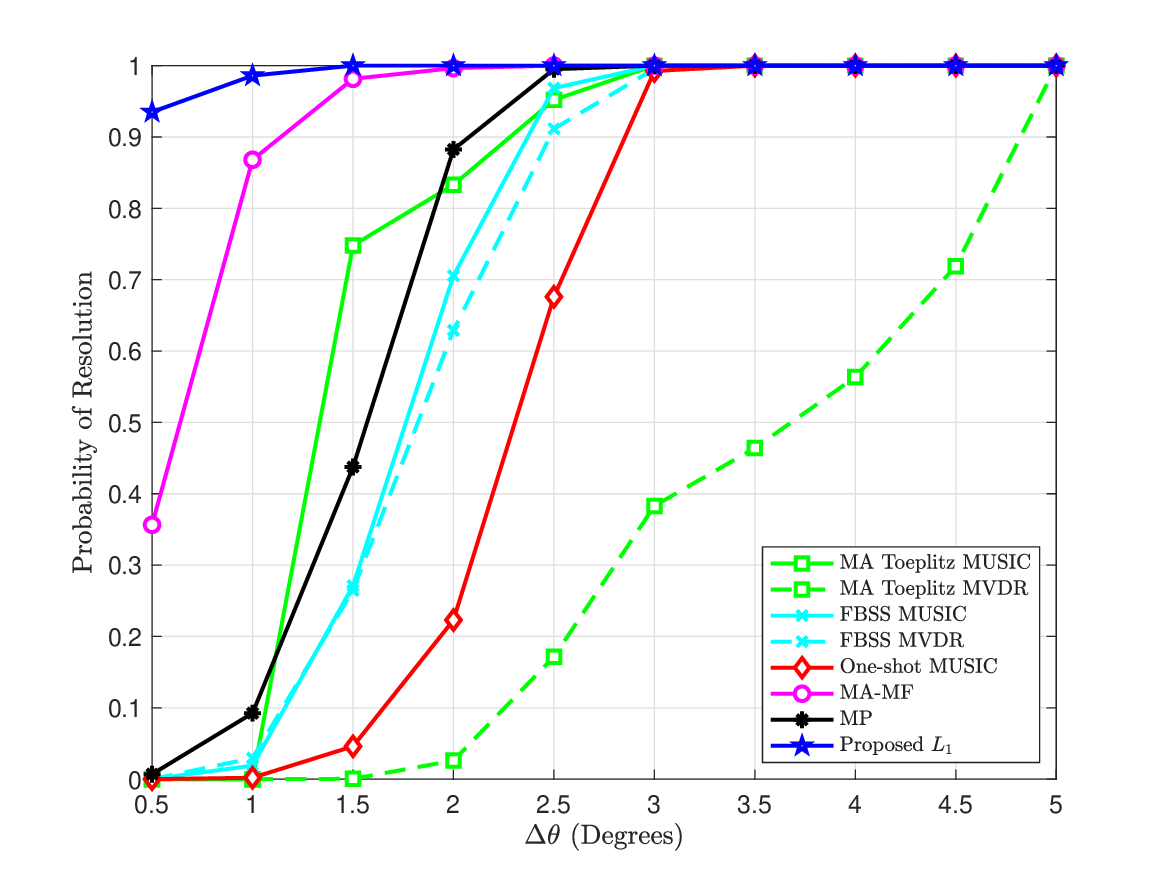}
        }
        \caption{Probability of resolution versus $\Delta\theta$ for $\text{SNR}=15$dB, impulse probability $p=0.1$: (a) $M=16$, and (b) $M=32$.}
        \label{fig:l1_prob_res_vs_dtheta_snr15_p0.1}
    \end{minipage}
    \centering
    \hfill
    \begin{minipage}[t]{0.48\textwidth}
        \centering
        \subfloat[]{
            \includegraphics[width=\textwidth]{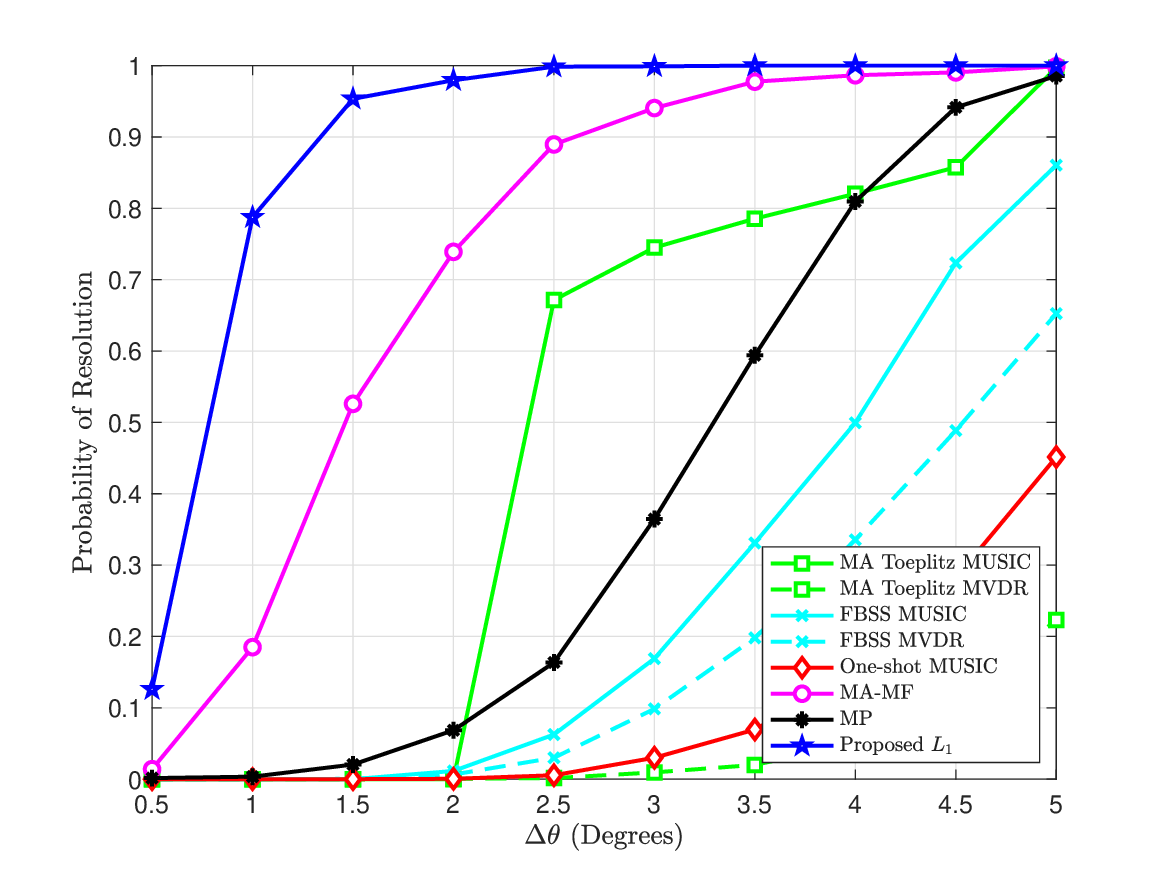}
        } \\
        \subfloat[]{
            \includegraphics[width=\textwidth]{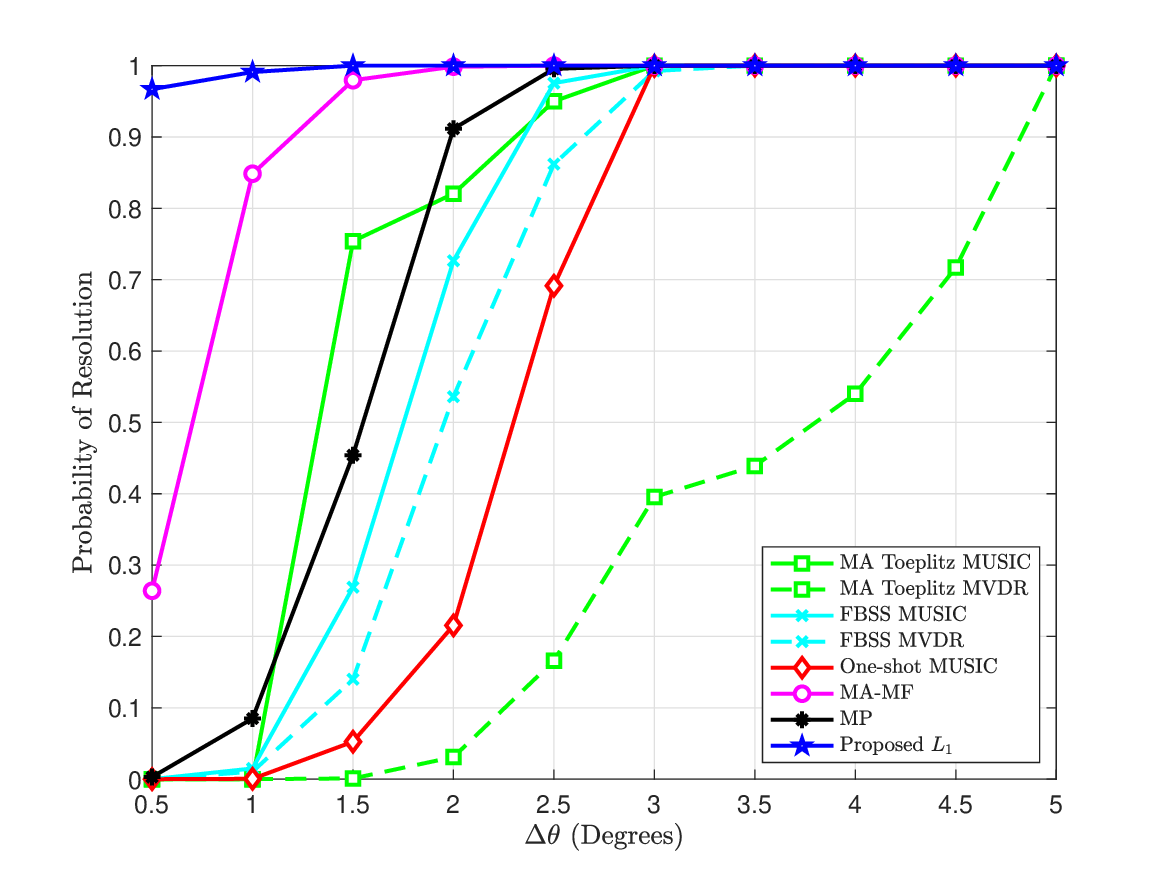}
        }
        \caption{Probability of resolution versus $\Delta\theta$ for $\text{SNR}=15$dB, impulse probability $p=0.25$: (a) $M=16$, and (b) $M=32$.}
        \label{fig:l1_prob_res_vs_dtheta_snr15_p0.25}
    \end{minipage}
\end{figure*}

Figs. \ref{fig:l1_prob_res_vs_dtheta_snr10_p0.1}-\ref{fig:l1_prob_res_vs_dtheta_snr15_p0.25}
complement the previous analysis by fixing the SNR at $10$dB and $15$dB
and examining the resolution probability as a function of the angular
separation $\Delta\theta$, for both impulse probabilities $p=0.1$ and
$p=0.25$. The proposed $L_1$-norm-based method consistently achieves the
highest resolution probability as $\Delta\theta$ increases, reaching
near-perfect resolution at separations where all competing approaches remain
far below acceptable performance levels. At SNR $= 10$dB with $M = 16$ and
$p = 0.1$ (Fig. \ref{fig:l1_prob_res_vs_dtheta_snr10_p0.1}(a)), the proposed
method already exceeds $80\%$ resolution probability at $\Delta\theta =
1.5^\circ$, while the closest competitor remains below $40\%$; for $M = 32$,
the proposed method resolves the two sources with near-certainty at the
same tested separation. At the higher SNR of $15$dB
(Figs. \ref{fig:l1_prob_res_vs_dtheta_snr15_p0.1}
and \ref{fig:l1_prob_res_vs_dtheta_snr15_p0.25}), the proposed method
achieves near perfect resolution across the entire range of $\Delta\theta$ for
$M=32$, regardless of impulse probability, whereas competing methods require
substantially larger separations to attain comparable probabilities. These
results jointly establish that the proposed method exhibits a significantly
smaller minimum resolvable angular separation than any competing framework
under matched conditions, a critical advantage in dense-source environments
corrupted by impulsive noise.

\section{Conclusions}

In this work, we addressed the problem of joint direction-of-arrival (DoA) estimation of multiple signals under limited sample support and hardware-constrained spatial sampling. We described a sliding subarray acquisition process, that we called Hankel sensing, which leads to a data matrix with Hankel-structured embedded signals component. Then, we developed 
a novel unified framework for direct $K$-signal DoA estimation by rank-$K$ Hankel-structured decomposition of the data matrix, under both the $L_2$ and $L_1$-norm formulation. The proposed $L_2$-norm estimator was shown to be maximum-likelihood optimal under white Gaussian noise. The $L_1$-norm estimator was shown to be maximum-likelihood optimal under i.i.d. Laplace noise.

Extensive simulation studies demonstrated that the proposed estimators achieve state-of-the-art resolution performance compared to competing approaches. In particular, the proposed methods require significantly lower SNR to resolve closely spaced sources and attain substantially higher probability of resolution, with the $L_1$-based estimator exhibiting pronounced robustness under impulsive noise conditions.

\appendix
\section*{Proof of Theorem \ref{l2_ml_equivalence}}
Defining the vectorized data observation
\begin{equation}
\mathbf{r}_v \triangleq \operatorname{vec}(\mathbf{X}) \in \mathbb{C}^{DW}
\end{equation}
and recalling the Hankel-structured manifold matrix $\mathbf{S}(\boldsymbol{\theta})$ in \eqref{s_theta}, the proposed estimator in \eqref{l2_vec_theta} can be rewritten as 
\begin{equation}
    \hat{\boldsymbol{\theta}}_{L_2}= \displaystyle \underset{\substack{ \boldsymbol{\theta} \in\left[-90^{\circ}, 90^{\circ}\right)^K}} {\operatorname{argmin}}\left\| \mathbf{r}_v- \mathbf{S}(\boldsymbol{\theta})\hat{\mathbf{c}}_{L_2}(\boldsymbol{\theta})\right\|_2.
    \label{Th1estimator}
\end{equation}
Substituting $\hat{\mathbf{c}}_{L_2}(\boldsymbol{\theta})$ from \eqref{c_l2} into the objective function yields
\begin{equation}
\left\|
\mathbf{r}_v-\mathbf{S}(\boldsymbol{\theta})\hat{\mathbf{c}}_{L_2}(\boldsymbol{\theta})
\right\|_2^2
=
\left\|
\mathbf{r}_v-\mathbf{P}_{\mathbf{S}(\boldsymbol{\theta})}\mathbf{r}_v
\right\|_2^2
\end{equation}
where
\begin{equation}
\mathbf{P}_{\mathbf{S}(\boldsymbol{\theta})}
=
\mathbf{S}(\boldsymbol{\theta})
\bigl(\mathbf{S}^H(\boldsymbol{\theta})\mathbf{S}(\boldsymbol{\theta})\bigr)^{-1}
\mathbf{S}^H(\boldsymbol{\theta})
\end{equation}
is the orthogonal projection matrix onto the column space of $\mathbf{S}(\boldsymbol{\theta})$.
Since $\mathbf{P}_{\mathbf{S}(\boldsymbol{\theta})}$ is Hermitian and idempotent,
\begin{equation}
\left\|
\mathbf{r}_v-\mathbf{P}_{\mathbf{S}(\boldsymbol{\theta})}\mathbf{r}_v
\right\|_2^2
=
\mathbf{r}_v^H\mathbf{r}_v
-
\mathbf{r}_v^H\mathbf{P}_{\mathbf{S}(\boldsymbol{\theta})}\mathbf{r}_v.
\end{equation}
The term $\mathbf{r}_v^H\mathbf{r}_v$ is independent of $\boldsymbol{\theta}$. Hence, solving (\ref{Th1estimator}) 
is equivalent to 
\begin{equation}
\hat{\boldsymbol{\theta}}_{L_2}
=
\underset{\boldsymbol{\theta}\in[-90^\circ,90^\circ)^K}{\operatorname{argmax}}
\mathbf{r}_v^H\mathbf{P}_{\mathbf{S}(\boldsymbol{\theta})}\mathbf{r}_v
\end{equation}
or
\begin{equation}
\hat{\boldsymbol{\theta}}_{L_2}
=
\underset{\boldsymbol{\theta}\in[-90^\circ,90^\circ)^K}{\operatorname{argmax}}
\mathbf{r}_v^H
\mathbf{S}(\boldsymbol{\theta})
\bigl(\mathbf{S}^H(\boldsymbol{\theta})\mathbf{S}(\boldsymbol{\theta})\bigr)^{-1}
\mathbf{S}^H(\boldsymbol{\theta})
\mathbf{r}_v.
\end{equation}
This is precisely the known deterministic maximum-likelihood DoA criterion for white Gaussian noise with the virtual received sample $\mathbf{r}_v$ replacing the conventional array snapshot and the Hankel-structured manifold matrix $\mathbf{S}(\boldsymbol{\theta})$ replacing the standard steering matrix \cite{526899,hacker2010single}. Therefore, the proposed estimator in \eqref{l2_vec_theta} is ML-optimal on the data matrix 
$\mathbf{X} \in \mathbb{C}^{DW}$ for white Gaussian noise.
\hfill $\blacksquare$

\section*{Proof of Theorem \ref{ml_laplace}}
Under the signal model in (\ref{signal_model_K_signals}) and sensing model in (\ref{sliding_sub}), (\ref{agg_meas_matrix}) the aggregate measurement matrix satisfies
\begin{equation}
\label{mat_model_app}
\mathbf{X} = \sum_{k=1}^{K} c_k \mathbf{s}_D\ \bigl(z(\theta_k)\bigr)\mathbf{s}_W\!\bigl(z(\theta_k)\bigr)^T + \mathbf{N}
\end{equation}
where $\mathbf{N}$ consists of i.i.d.\ circularly symmetric complex Laplace entries. Define the vectorized data observation
\begin{equation}
\mathbf{r}_v \triangleq \operatorname{vec}(\mathbf{X}) \in \mathbb{C}^{DW}
\end{equation}
and recall the Hankel-structured manifold matrix $\mathbf{S}(\boldsymbol{\theta})$ in \eqref{s_theta}. Then, by vectorization of \eqref{mat_model_app} we have the equivalent virtual measurement model
\begin{equation}
\label{virtual_model_app}
\mathbf{r}_v = \mathbf{S}(\boldsymbol{\theta})\mathbf{c} + \mathbf{n}
\end{equation}
where $\mathbf{n} = \operatorname{vec}(\mathbf{N})$.

Under the i.i.d.\ circularly symmetric complex Laplace assumption the probability density function of the measurement vector $\mathbf{r}_v$ is 
\begin{equation}
f(\mathbf{r}_v \mid \boldsymbol{\theta},\mathbf{c}) =
\left(\frac{\lambda^2}{2\pi}\right)^{DW}
\exp\left( -\lambda \|\mathbf{r}_v-\mathbf{S}(\boldsymbol{\theta})\mathbf{c}\|_1 \right), 
\end{equation}
for some $\lambda > 0$.
Therefore, the negative log-likelihood is
\begin{equation} 
-\log f(\mathbf{r}_v \mid \boldsymbol{\theta},\mathbf{c}) = -DW\log\ \left(\frac{\lambda^2}{2\pi}\right)
+ \lambda \|\mathbf{r}_v-\mathbf{S} (\boldsymbol{\theta})\mathbf{c}\|_1.
\end{equation}
Since the first term is independent of $(\boldsymbol{\theta},\mathbf{c})$ and $\lambda>0$, maximizing the likelihood is equivalent to minimizing 
\begin{equation}
    \|\mathbf{r}_v-\mathbf{S}(\boldsymbol{\theta})\mathbf{c}\|_1.
\end{equation}
Thus,
\begin{equation}
    (\hat{\mathbf{c}}_{L_1},\hat{\boldsymbol{\theta}}_{L_1}) \in \displaystyle \underset{\substack{\mathbf{c} \in \mathbb{C}^{K}, \\ \boldsymbol{\theta} \in\left[-90^{\circ}, 90^{\circ}\right)^K}} {\operatorname{argmin}}\left\| \mathbf{r}_v- \mathbf{S}(\boldsymbol{\theta})\mathbf{c}\right\|_1.
\end{equation}

For each fixed $\boldsymbol{\theta}$, minimization with respect to $\mathbf{c}$ gives $\hat{c}_{L_1}(z(\theta))$ as defined in (\ref{lad}) and substitution into the objective yields 
\begin{equation}
    \hat{\boldsymbol{\theta}}_{L_1}= \displaystyle \underset{\substack{ \boldsymbol{\theta} \in\left[-90^{\circ}, 90^{\circ}\right)^K}} {\operatorname{argmin}}\left\| \mathbf{r}_v- \mathbf{S}(\boldsymbol{\theta})\hat{\mathbf{c}}_{L_1}(\boldsymbol{\theta})\right\|_1
\end{equation}
which is exactly the proposed estimator in \eqref{l1_vec_theta} with $\mathbf{r}_v = \mathrm{vec}(\mathbf{X})$. 
\hfill $\blacksquare$

\bibliographystyle{IEEEtran}
\bibliography{IEEEabrv,bibliography}

@article{orfanidis2026hankel,
  author  = {G. I. Orfanidis and D. A. Pados and G. Sklivanitis and E. S. Bentley},
  title   = {Hankel and {T}oeplitz Rank-1 Decomposition of Arbitrary Matrices with Applications to Few-Shot Signal Direction-of-Arrival Estimation},
  journal = {IEEE Trans. Signal Process.},
  note    = {submitted, Apr. 2026},
}

@book{boyd2004convex,
  author    = {Stephen Boyd and Lieven Vandenberghe},
  title     = {Convex Optimization},
  publisher = {Cambridge Univ. Press},
  address   = {Cambridge, U.K.},
  year      = {2004}
}

@article{schlossmacher1973,
  author  = {E. J. Schlossmacher},
  title   = {An Iterative Technique for Absolute Deviations Curve Fitting},
  journal = {J. Amer. Stat. Assoc.},
  volume  = {68},
  number  = {344},
  pages   = {857--859},
  month   = dec,
  year    = {1973}
}

@article{wesolowsky1981,
  author  = {G. O. Wesolowsky},
  title   = {A New Descent Algorithm for the Least Absolute Value Regression Problem},
  journal = {Commun. Statist. Simul. Comput.},
  volume  = {10},
  number  = {5},
  pages   = {479--491},
  year    = {1980}, 
  month = jul
}

@article{57542,
  author  = {Petre Stoica and K. C. Sharman},
  title   = {Maximum Likelihood Methods for Direction-of-Arrival Estimation},
  journal = {IEEE Trans. Acoust., Speech, Signal Process.},
  volume  = {38},
  number  = {7},
  pages   = {1132--1143},
  month   = jul,
  year    = {1990}
}

@article{scheweppe,
  author  = {Schweppe, F.F},
  title   = {Sensor Array Data Processing for Multiple-Signal Sources},
  journal = {IEEE Trans. Inf. Theory},
  volume  = {14},
  number  = {2},
  pages   = {294--305},
  month   = mar,
  year    = {1968}
}

@article{17564,
  author  = {Petre Stoica and Arye Nehorai},
  title   = {{MUSIC}, Maximum Likelihood, and {Cram{\'e}r--Rao} Bound},
  journal = {IEEE Trans. Acoust., Speech, Signal Process.},
  volume  = {37},
  number  = {5},
  pages   = {720--741},
  month   = may,
  year    = {1989},
}

@article{348129,
  author  = {Mats Viberg and Bj{\"o}rn Ottersten and Arye Nehorai},
  title   = {Performance Analysis of Direction Finding with Large Arrays and Finite Data},
  journal = {IEEE Trans. Signal Process.},
  volume  = {43},
  number  = {2},
  pages   = {469--477},
  month   = feb,
  year    = {1995},
}

@article{7494995,
  author  = {Yang Cao and Tao Jiang and Zhu Han},
  title   = {A Survey of Emerging {M2M} Systems: Context, Task, and Objective},
  journal = {IEEE Internet Things J.},
  volume  = {3},
  number  = {6},
  pages   = {1246--1258},
  month   = jun,
  year    = {2016},
}

@article{8879484,
  author  = {Lalit Chettri and Rabindranath Bera},
  title   = {A Comprehensive Survey on Internet of Things ({IoT}) toward 5G Wireless Systems},
  journal = {IEEE Internet Things J.},
  volume  = {7},
  number  = {1},
  pages   = {16--32},
  month   = jan,
  year    = {2020},
}

@article{9214481,
  author  = {Ziqi Chen and David Smith},
  title   = {{mmWave} {M2M} Networks: Improving Delay Performance of Relaying},
  journal = {IEEE Trans. Wireless Commun.},
  volume  = {20},
  number  = {1},
  pages   = {577--589},
  month   = jan,
  year    = {2021},
}

@article{9362210,
  author  = {Riheng Wu and Mei Wang and Zhenhai Zhang},
  title   = {Computationally Efficient {DOA} and Carrier Estimation for Coherent Signal Using Single Snapshot and Its Time-Delay Replications},
  journal = {IEEE Trans. Aerosp. Electron. Syst.},
  volume  = {57},
  number  = {4},
  pages   = {2469--2480},
  month   = aug,
  year    = {2021},
}

@article{9535488,
  author  = {Chenglong Li and Sibren De Bast and Emmeric Tanghe and Sofie Pollin and Wout Joseph},
  title   = {Toward Fine-Grained Indoor Localization Based on Massive {MIMO}-{OFDM} System: Experiment and Analysis},
  journal = {IEEE Sensors J.},
  volume  = {22},
  number  = {6},
  pages   = {5318--5328},
  month   = mar,
  year    = {2022},
}

@article{9965430,
  author  = {Teng Ma and Yue Xiao and Xia Lei},
  title   = {Channel Reconstruction-Aided {MUSIC} Algorithms for Joint {AoA} \& {AoD} Estimation in {MIMO} Systems},
  journal = {IEEE Wireless Commun. Lett.},
  volume  = {12},
  number  = {2},
  pages   = {322--326},
  month   = feb,
  year    = {2023},
}

@article{7744507,
  author  = {Qing Shen and Wei Liu and Wei Cui and Siliang Wu},
  title   = {Underdetermined {DOA} Estimation under the Compressive Sensing Framework: A Review},
  journal = {IEEE Access},
  volume  = {4},
  pages   = {8865--8878},
  month = nov,
  year    = {2016}}

@article{1468495,
  author  = {D. Malioutov and M. {\c{C}}etin and A. S. Willsky},
  title   = {A Sparse Signal Reconstruction Perspective for Source Localization with Sensor Arrays},
  journal = {IEEE Trans. Signal Process.},
  volume  = {53},
  number  = {8},
  pages   = {3010--3022},
  month   = aug,
  year    = {2005},
}

@article{4663911,
  author  = {Hosein Mohimani and Massoud Babaie-Zadeh and Christian Jutten},
  title   = {A Fast Approach for Overcomplete Sparse Decomposition Based on Smoothed {$\ell_0$} Norm},
  journal = {IEEE Trans. Signal Process.},
  volume  = {57},
  number  = {1},
  pages   = {289--301},
  month   = jan,
  year    = {2009},}

@article{5599897,
  author  = {Petre Stoica and Prabhu Babu and Jian Li},
  title   = {New Method of Sparse Parameter Estimation in Separable Models and Its Use for Spectral Analysis of Irregularly Sampled Data},
  journal = {IEEE Trans. Signal Process.},
  volume  = {59},
  number  = {1},
  pages   = {35--47},
  month   = jan,
  year    = {2011},
}

@article{5417172,
  author  = {Tarik Yardibi and Jian Li and Petre Stoica and Ming Xue and Arthur B. Baggeroer},
  title   = {Source Localization and Sensing: A Nonparametric Iterative Adaptive Approach Based on Weighted Least Squares},
  journal = {IEEE Trans. Aerosp. Electron. Syst.},
  volume  = {46},
  number  = {1},
  pages   = {425--443},
  month   = jan,
  year    = {2010},
}

@article{fortunati2014single,
  author  = {Stefano Fortunati and Raffaele Grasso and Fulvio Gini and Maria S. Greco and Kevin LePage},
  title   = {Single-Snapshot {DOA} Estimation by Using Compressed Sensing},
  journal = {EURASIP J. Adv. Signal Process.},
  volume  = {2014},
  pages   = {1--17},
  month = jul, 
  year    = {2014}}

@article{Belloni_2011,
  author  = {A. Belloni and V. Chernozhukov and L. Wang},
  title   = {Square-Root Lasso: Pivotal Recovery of Sparse Signals via Conic Programming},
  journal = {Biometrika},
  volume  = {98},
  number  = {4},
  pages   = {791--806},
  month   = nov,
  year    = {2011}
}

@article{babu2014connection,
  author  = {Prabhu Babu and Petre Stoica},
  title   = {Connection Between {SPICE} and Square-Root {LASSO} for Sparse Parameter Estimation},
  journal = {Signal Process.},
  volume  = {95},
  pages   = {10--14},
  year    = {2014}, 
  month = feb
}

@article{6553252,
  author  = {Cristian R. Rojas and Dimitrios Katselis and H{\aa}kan Hjalmarsson},
  title   = {A Note on the {SPICE} Method},
  journal = {IEEE Trans. Signal Process.},
  volume  = {61},
  number  = {18},
  pages   = {4545--4551},
  month   = sep,
  year    = {2013}
}

@article{gorodnitsky1997sparse,
  author  = {Irina F. Gorodnitsky and Bhaskar D. Rao},
  title   = {Sparse Signal Reconstruction from Limited Data Using {FOCUSS}: A Re-Weighted Minimum Norm Algorithm},
  journal = {IEEE Trans. Signal Process.},
  volume  = {45},
  number  = {3},
  pages   = {600--616},
  month   = mar,
  year    = {1997}
}

@article{rao2003subset,
  author  = {Bhaskar D. Rao and Kjersti Engan and Shane F. Cotter and Jason Palmer and Kenneth Kreutz-Delgado},
  title   = {Subset Selection in Noise Based on Diversity Measure Minimization},
  journal = {IEEE Trans. Signal Process.},
  volume  = {51},
  number  = {3},
  pages   = {760--770},
  month   = mar,
  year    = {2003}
}

@article{tan2010sparse,
  author  = {Xing Tan and William Roberts and Jian Li and Petre Stoica},
  title   = {Sparse Learning via Iterative Minimization with Application to {MIMO} Radar Imaging},
  journal = {IEEE Trans. Signal Process.},
  volume  = {59},
  number  = {3},
  pages   = {1088--1101},
  month   = mar,
  year    = {2011}
}

@incollection{yang2018sparse,
  author    = {Zai Yang and Jian Li and Petre Stoica and Lihua Xie},
  title     = {Sparse Methods for Direction-of-Arrival Estimation},
  booktitle = {Academic Press Library in Signal Processing, Volume 7},
  publisher = {Academic Press},
  pages     = {509--581},
  year      = {2018}
}

@article{6867345,
  author  = {Yuxin Chen and Yuejie Chi},
  title   = {Robust Spectral Compressed Sensing via Structured Matrix Completion},
  journal = {IEEE Trans. Inf. Theory},
  volume  = {60},
  number  = {10},
  pages   = {6576--6601},
  month   = oct,
  year    = {2014}
}

@article{6576276,
  author  = {Gongguo Tang and Badri Narayan Bhaskar and Parikshit Shah and Benjamin Recht},
  title   = {Compressed Sensing Off the Grid},
  journal = {IEEE Trans. Inf. Theory},
  volume  = {59},
  number  = {11},
  pages   = {7465--7490},
  month   = nov,
  year    = {2013}
}

@inproceedings{10056957,
  author    = {Nir Shlezinger and Jay Whang and Yonina C. Eldar and Alexandros G. Dimakis},
  title     = {Model-Based Deep Learning: Key Approaches and Design Guidelines},
  booktitle = {Proc. IEEE Data Sci. Learn. Workshop (DSLW)},
  year      = {2021},
  pages     = {1--6},
  address   = {Toronto, ON, Canada},
  month     = jun
}

@inproceedings{9747692,
  author    = {Marcio L. Lima de Oliveira and Marco J. G. Bekooij},
  title     = {{Deep-MLE}: Fusion between a Neural Network and {MLE} for a Single Snapshot {DOA} Estimation},
  booktitle = {Proc. IEEE Int. Conf. Acoust., Speech, Signal Process. (ICASSP)},
  year      = {2022},
  pages     = {3673--3677},
  address   = {Singapore},
  month     = may
}

@inproceedings{8726554,
  author    = {Jonas Fuchs and Robert Weigel and Markus Gardill},
  title     = {Single-Snapshot Direction-of-Arrival Estimation of Multiple Targets Using a Multi-Layer Perceptron},
  booktitle = {Proc. IEEE MTT-S Int. Conf. Microw. Intell. Mobility (ICMIM)},
  year      = {2019},
  pages     = {1--4},
  address   = {Detroit, MI, USA},
  month     = apr
}

@article{120069,
  author  = {S. Jha and T. Durrani},
  title   = {Direction of Arrival Estimation Using Artificial Neural Networks},
  journal = {IEEE Trans. Syst., Man, Cybern.},
  volume  = {21},
  number  = {5},
  pages   = {1192--1201},
  month   = sep,
  year    = {1991}
}

@inproceedings{1169330,
  author    = {R. Rastogi and P. Gupta and R. Kumaresan},
  title     = {Array Signal Processing with Interconnected Neuron-Like Elements},
  booktitle = {Proc. IEEE Int. Conf. Acoust., Speech, Signal Process. (ICASSP)},
  year      = {1987},
  pages     = {2328--2331},
  address   = {Dallas, TX, USA},
  month     = apr
}

@inproceedings{197061,
  author    = {D. Goryn and M. Kaveh},
  title     = {Neural Networks for Narrowband and Wideband Direction Finding},
  booktitle = {Proc. IEEE Int. Conf. Acoust., Speech, Signal Process. (ICASSP)},
  year      = {1988},
  pages     = {2164--2167},
  address   = {New York, NY, USA},
  month     = apr
}

@inproceedings{8682604,
  author    = {Oded Bialer and Noa Garnett and Tom Tirer},
  title     = {Performance Advantages of Deep Neural Networks for Angle of Arrival Estimation},
  booktitle = {Proc. IEEE Int. Conf. Acoust., Speech, Signal Process. (ICASSP)},
  year      = {2019},
  pages     = {3907--3911},
  address   = {Brighton, U.K.},
  month     = may
}

@article{10348517,
  author  = {Ruxin Zheng and Shunqiao Sun and Hongshan Liu and Honglei Chen and Jian Li},
  title   = {Interpretable and Efficient Beamforming-Based Deep Learning for Single Snapshot {DOA} Estimation},
  journal = {IEEE Sensors J.},
  year    = {2023},
  month = dec, 
  note    = {Early Access}
}

@article{hacker2010single,
  author  = {Patrick H{\"a}cker and B. Yang},
  title   = {Single Snapshot {DOA} Estimation},
  journal = {Adv. Radio Sci.},
  volume  = {8},
  pages   = {251--256},
  month = sep,
  year    = {2010}
}

@book{birkes2011alternative,
  author    = {David Birkes and Yadolah Dodge},
  title     = {Alternative Methods of Regression},
  publisher = {Wiley},
  address   = {New York, NY, USA},
  year      = {1993}
}

@article{bloomfield1980least,
  author  = {Peter Bloomfield and William Steiger},
  title   = {Least Absolute Deviations Curve-Fitting},
  journal = {SIAM J. Sci. Stat. Comput.},
  volume  = {1},
  number  = {2},
  pages   = {290--301},
  year    = {1980}
}

@book{golub1996matrix,
  author    = {Gene H. Golub and Charles F. Van Loan},
  title     = {Matrix Computations},
  edition   = {3rd},
  publisher = {The Johns Hopkins University Press},
  address   = {Baltimore, MD, USA},
  year      = {1996}
}

@book{van2002optimum,
  title={Optimum array processing: Part IV of detection, estimation, and modulation theory},
  author={Van Trees, Harry L},
  year={2002},
  month=apr,
  publisher={John Wiley \& Sons}
}

@inproceedings{orfanidis2022time,
  author    = {Georgios I. Orfanidis and Dimitrios A. Pados and George Sklivanitis},
  title     = {Time-Series Analysis with Small and Faulty Data: {L1}-Norm Decompositions of Hankel Matrices},
  booktitle = {Proc. SPIE Big Data IV: Learning, Analytics, and Applications},
  volume    = {12097},
  pages     = {97--104},
  address   = {Orlando, FL, USA},
  month     = may,
  year      = {2022},
}

@inproceedings{10051870,
  author    = {Georgios I. Orfanidis and Dimitrios A. Pados and George Sklivanitis and Elizabeth S. Bentley and Joseph Suprenant and Michael J. Medley},
  title     = {Single-Sample Direction-of-Arrival Estimation by {Hankel}-Matrix Decompositions},
  booktitle = {Proc. 56th Asilomar Conf. Signals, Syst., Comput.},
  year      = {2022},
  pages     = {1026--1030},
  address   = {Pacific Grove, CA, USA},
  month     = nov,
}

@inproceedings{10096647,
  author    = {Stepan Mazokha and Sanaz Naderi and Georgios I. Orfanidis and George Sklivanitis and Dimitrios A. Pados and Jason O. Hallstrom},
  title     = {Single-Sample Direction-of-Arrival Estimation for Fast and Robust 3D Localization with Real Measurements from a Massive {MIMO} System},
  booktitle = {Proc. IEEE Int. Conf. Acoust., Speech, Signal Process. (ICASSP)},
  year      = {2023},
  pages     = {1--5},
  address   = {Rhodes Island, Greece},
  month     = jun,
}

@article{knirsch2021optimal,
  author  = {Hanna Knirsch and Markus Petz and Gerlind Plonka},
  title   = {Optimal Rank-1 {Hankel} Approximation of Matrices: Frobenius Norm and Spectral Norm and {Cadzow}'s Algorithm},
  journal = {Linear Algebra and its Appl.},
  volume  = {629},
  pages   = {1--39},
  year    = {2021},
  month = nov
}

@article{526899,
  author  = {Hamid Krim and Mats Viberg},
  title   = {Two Decades of Array Signal Processing Research: The Parametric Approach},
  journal = {IEEE Signal Process. Mag.},
  volume  = {13},
  number  = {4},
  pages   = {67--94},
  month   = jul,
  year    = {1996},
}

@article{622504,
  author  = {Lalit C. Godara},
  title   = {Application of Antenna Arrays to Mobile Communications. {II}. Beam-Forming and Direction-of-Arrival Considerations},
  journal = {Proc. IEEE},
  volume  = {85},
  number  = {8},
  pages   = {1195--1245},
  month   = aug,
  year    = {1997},
}

@article{7815340,
  author  = {Ulrik Nielsen and Jie-Bang Yan and Sivaprasad Gogineni and J{\o}rgen Dall},
  title   = {Direction-of-Arrival Analysis of Airborne Ice Depth Sounder Data},
  journal = {IEEE Trans. Geosci. Remote Sens.},
  volume  = {55},
  number  = {4},
  pages   = {2239--2249},
  month   = apr,
  year    = {2017},
}

@article{9219157,
  author  = {Jiurui Zhao and Yingwei Tian and Biyang Wen and Zhen Tian},
  title   = {Coherent {DOA} Estimation in Sea Surface Observation with Direction-Finding {HF} Radar},
  journal = {IEEE Trans. Geosci. Remote Sens.},
  volume  = {59},
  number  = {8},
  pages   = {6651--6661},
  month   = aug,
  year    = {2021},
}

@inproceedings{9725025,
  author    = {Pengfei Li and Yubo Tian},
  title     = {{DOA} Estimation of Underwater Acoustic Signals Based on Deep Learning},
  booktitle = {Proc. 2nd Int. Seminar Artif. Intell., Netw. Inf. Technol. (AINIT)},
  year      = {2021},
  pages     = {221--225},
  address   = {Shanghai, China},
  month     = oct,
}

@article{9722876,
  author  = {Sushil Kumar Joshi and Stefan V. Baumgartner and Andr{\'e} Barros Cardoso da Silva and Gerhard Krieger},
  title   = {Direction-of-Arrival Angle and Position Estimation for Extended Targets Using Multichannel Airborne Radar Data},
  journal = {IEEE Geosci. Remote Sens. Lett.},
  volume  = {19},
  pages   = {1--5},
  year    = {2022},
  month = feb
}

@inproceedings{9148550,
  author    = {Matilde Boschiero and Marco Giordani and Michele Polese and Michele Zorzi},
  title     = {Coverage Analysis of {UAV}s in Millimeter Wave Networks: A Stochastic Geometry Approach},
  booktitle = {Proc. Int. Wireless Commun. Mobile Comput. Conf. (IWCMC)},
  year      = {2020},
  pages     = {351--357},
  address   = {Limassol, Cyprus},
  month     = jun,
}

@article{9726790,
  author  = {Mingsheng Yin and Akshaj Kumar Veldanda and Amee Trivedi and Jeff Zhang and Kai Pfeiffer and Yaqi Hu and Siddharth Garg and Elza Erkip and Ludovic Righetti and Sundeep Rangan},
  title   = {Millimeter Wave Wireless Assisted Robot Navigation with Link State Classification},
  journal = {IEEE Open J. Commun. Soc.},
  volume  = {3},
  pages   = {493--507},
  year    = {2022},
  month = mar
}

@inproceedings{10160524,
  author    = {Kai Pfeiffer and Yuze Jia and Mingsheng Yin and Akshaj Kumar Veldanda and Yaqi Hu and Amee Trivedi and Jeff Zhang and Siddharth Garg and Elza Erkip and Sundeep Rangan and Ludovic Righetti},
  title     = {Path Planning Under Uncertainty to Localize {mmWave} Sources},
  booktitle = {Proc. IEEE Int. Conf. Robot. Autom. (ICRA)},
  year      = {2023},
  pages     = {3461--3467},
  address   = {London, U.K.},
  month     = jun,
}

@article{10018231,
  author  = {Bisma Amjad and Qasim Zeeshan Ahmed and Pavlos I. Lazaridis and Maryam Hafeez and Faheem A. Khan and Zaharias D. Zaharis},
  title   = {{Radio SLAM}: A Review on Radio-Based Simultaneous Localization and Mapping},
  journal = {IEEE Access},
  volume  = {11},
  pages   = {9260--9278},
  year    = {2023},
}

@article{32276,
  author  = {R. Roy and Thomas Kailath},
  title   = {{ESPRIT}—Estimation of Signal Parameters via Rotational Invariance Techniques},
  journal = {IEEE Trans. Acoust., Speech, Signal Process.},
  volume  = {37},
  number  = {7},
  pages   = {984--995},
  month   = jul,
  year    = {1989},
}

@article{80966,
  author  = {Viberg, M. and Ottersten, B.},
  title   = {Sensor Array Processing Based on Subspace Fitting},
  journal = {IEEE Trans. Signal Process.},
  volume  = {39},
  number  = {5},
  pages   = {1110--1121},
  month   = may,
  year    = {1991}
}

@article{4063549,
  author  = {Rohan Grover and Dimitrios A. Pados and Michael J. Medley},
  title   = {Subspace Direction Finding with an Auxiliary-Vector Basis},
  journal = {IEEE Trans. Signal Process.},
  volume  = {55},
  number  = {2},
  pages   = {758--763},
  month   = jan,
  year    = {2007},
}

@article{9516894,
  author  = {Ivan Podkurkov and Gabriel Seidl and Liana Khamidullina and Adel Nadeev and Martin Haardt},
  title   = {Tensor-Based Near-Field Localization Using Massive Antenna Arrays},
  journal = {IEEE Trans. Signal Process.},
  volume  = {69},
  pages   = {5830--5845},
  year    = {2021},
  month = aug
}

@article{hua2002matrix,
  author={Hua, Y. and Sarkar, T.K.},
  title   = {Matrix Pencil Method for Estimating Parameters of Exponentially Damped/Undamped Sinusoids in Noise},
  journal = {IEEE Trans. Acoust., Speech, Signal Process.},
  volume  = {38},
  number  = {5},
  pages   = {814--824},
  month   = may,
  year    = {1990}
}

@phdthesis{adve1996elimination,
  author = {Raviraj Sadanand Adve},
  title  = {Elimination of the Effects of Mutual Coupling in Adaptive Thin Wire Antennas},
  school = {Syracuse University},
  address= {Syracuse, NY, USA},
  year   = {1996}
}

@article{liao2016music,
  author  = {Wenjing Liao and Albert Fannjiang},
  title   = {{MUSIC} for Single-Snapshot Spectral Estimation: Stability and Super-Resolution},
  journal = {Appl. Comput. Harmon. Anal.},
  volume  = {40},
  number  = {1},
  pages   = {33--67},
  year    = {2016},
  month = jan
}

@inproceedings{9564893,
  author    = {Chengliang Liu and Weike Feng and Hongbing Li and Hangui Zhu},
  title     = {Single Snapshot {DOA} Estimation Based on Spatial Smoothing {MUSIC} and {CNN}},
  booktitle = {Proc. IEEE Int. Conf. Signal Process., Commun. Comput. (ICSPCC)},
  year      = {2021},
  pages     = {1--5},
  address   = {Xi'an, China},
  month     = aug,
}

@inproceedings{degen2017single,
  author    = {Christoph Degen},
  title     = {On Single Snapshot Direction-of-Arrival Estimation},
  booktitle = {Proc. IEEE Int. Conf. Wireless Space Extreme Environ. (WiSEE)},
  year      = {2017},
  pages     = {92--97},
  address   = {Montr{\'e}al, QC, Canada},
  month     = oct,
}

@book{marple2019digital,
 author    = {Marple, Jr.,S. Lawrence},
  title     = {Digital Spectral Analysis},
  publisher = {Dover Publications},
  address   = {Mineola, NY, USA},
  year      = {2019},
  month = mar
}

@article{1164649,
  author={Tie-Jun Shan and Wax, M. and Kailath, T.},
  title   = {On Spatial Smoothing for Direction-of-Arrival Estimation of Coherent Signals},
  journal = {IEEE Trans. Acoust., Speech, Signal Process.},
  volume  = {33},
  number  = {4},
  pages   = {806--811},
  month   = aug,
  year    = {1985},
}

@article{1143830,
  author={Schmidt, R.},
  title   = {Multiple Emitter Location and Signal Parameter Estimation},
  journal = {IEEE Trans. Antennas Propag.},
  volume  = {34},
  number  = {3},
  pages   = {276--280},
  month   = mar,
  year    = {1986},
}
\end{document}